\pdfoutput=1

\documentclass[11pt]{article}

\usepackage{ACL2023}

\usepackage{times}
\usepackage{latexsym}

\usepackage[T1]{fontenc}

\usepackage[utf8]{inputenc}

\usepackage{microtype}

\usepackage{inconsolata}
\usepackage[utf8]{inputenc} 
\usepackage[T1]{fontenc}    
\usepackage{hyperref}       
\usepackage{url}            
\usepackage{booktabs}       
\usepackage{amsfonts}       
\usepackage{nicefrac}       
\usepackage{microtype}      
\usepackage{soul}
\usepackage{subcaption}
\usepackage{multicol}
\usepackage{graphicx}
\usepackage{wrapfig}
\usepackage{multirow}
\usepackage{makecell}
\usepackage{xspace}
\usepackage{amsmath}
\usepackage{listings}
\usepackage{algorithm}
\usepackage{amsfonts,amssymb}
\usepackage{mathrsfs}
\usepackage{subcaption}
\usepackage{natbib}
\usepackage{graphicx}       
\usepackage{cleveref}
\usepackage{minitoc}
\usepackage{color}
\usepackage{xcolor}
\usepackage{listings}
\usepackage{multirow}
\usepackage[utf8]{inputenc} 
\usepackage[T1]{fontenc}    
\usepackage{hyperref}       
\usepackage{url}            
\usepackage{booktabs}       
\usepackage{amsfonts}       
\usepackage{nicefrac}       
\usepackage{microtype}      
\usepackage{xcolor}         
\usepackage{algorithm}
\usepackage{enumitem}
\usepackage{amsmath}
\usepackage{multirow}
\usepackage{array}
\usepackage{colortbl}
\usepackage{graphicx}
\usepackage{tabularx}
\usepackage{color, soul}
\usepackage{makecell}
\usepackage{amssymb}
\usepackage{arydshln}
\usepackage{geometry}
\usepackage{caption}
\usepackage{xspace}
\usepackage[utf8]{inputenc}
\usepackage{CJKutf8}
\usepackage{arydshln} 
\usepackage{algorithm}
\usepackage{algorithmicx}
\usepackage{tikz}
\usepackage{asymptote}
\usepackage{tikz}
\usepackage{amsmath}
\usepackage{algorithm}
\usepackage{ulem}
\usepackage{algpseudocode}
\usepackage{booktabs}
\usepackage{multirow}
\usepackage{microtype}
\usepackage{tabularx} 
\usepackage{longtable}
\usepackage{xspace}
\usepackage{enumitem}
\usepackage{graphicx}
\usepackage{lscape}
\usepackage{bbding}
\usepackage[para,online,flushleft]{threeparttable}
\usepackage{xspace}
\newcommand{\vpara}[1]{\noindent\textbf{#1}\xspace}

\useunder{\uline}{\ul}{}

\newcommand{\hide}[1]{} 
\usepackage{multirow}

\title{AndroidLab: Training and Systematic Benchmarking of Android Autonomous Agents}

\author{
Yifan Xu$^{1*}$, Xiao Liu$^{1*}$, Xueqiao Sun$^{1}$, Siyi Cheng$^{2\dagger}$, Hao Yu$^{1}$, Hanyu Lai$^{1}$,\\
\bf{Shudan Zhang$^{1}$, Dan Zhang$^{1}$, Jie Tang$^{1}$, Yuxiao Dong$^{1}$}\\ \\
\textsuperscript{1}Tsinghua University \quad
\textsuperscript{2}Peking University
}

\begin{document}

\maketitle

\renewcommand{\thefootnote}{\fnsymbol{footnote}}
    \footnotetext[1]{Yifan and Xiao contributed equally. Emails: \texttt{xu-yf23@mails.tsinghua.edu.cn},\texttt{shawliu9@gmail.com}}
    \footnotetext[2]{Work done when these authors visited Tsinghua University.}
\renewcommand{\thefootnote}{\arabic{footnote}}

\begin{abstract}
    Autonomous agents have become increasingly important for interacting with the real world. 
    Android agents, in particular, have been recently a frequently-mentioned interaction method. 
    However, existing studies for training and evaluating Android agents lack systematic research on both open-source and closed-source models. 
    In this work, we propose \textsc{AndroidLab} as a systematic Android agent framework. It includes an operation environment with different modalities, action space, and a reproducible benchmark. 
    It supports both large language models (LLMs) and multimodal models (LMMs) in the same action space. 
    \textsc{AndroidLab} benchmark includes predefined Android virtual devices and 138 tasks across nine apps built on these devices. 
    By using the \textsc{AndroidLab} environment, we develop an Android Instruction dataset and train six open-source LLMs and LMMs, lifting the average success rates from 4.59\% to 21.50\% for LLMs and from 1.93\% to 13.28\% for LMMs. 
    \textsc{AndroidLab} is open-sourced and publicly available at \url{https://github.com/THUDM/Android-Lab}.
\end{abstract}

\hide{Autonomous agents have become increasingly important for interacting with the real world. Android agents, in particular, have been a frequently mentioned interaction method in recent research. However, current benchmarks for evaluating Android agents still require further development. They lack a good combination of reproducibility and task difficulty. In this work, we propose \textsc{AndroidLab} as a systematic Android agent framework. It includes an operation environment with operation modes, action space, and a reproducible benchmark---\bench. It supports both \textit{text-based} and \textit{multi-modal} models in the same action space. \bench includes predefined Android virtual images and 138 tasks across nine apps built on these images. Using this environment, we created the Android Instruction dataset and trained six open-source LLMs and LMMs, achieving an increase in average success rates from 4.56\% to 16.43\% for LLMs and 1.93\% to 11.83\% for LMMs. Our benchmark and code are open-sourced and available at \url{https://github.com/THUDM/Android-Lab}.}

\section{Introduction}

Developing autonomous agents to execute human instructions within mobile operating systems has long been a goal for researchers~\cite{burns2021mobile-motif,yang2023appagent,wang2023enabling,hong2023cogagent,rawles2023aitw,li-etal-2020-mapping-pixelhelp,romao2019robotic,rai2019robotic}. 
Recently, a significant line of research has focused on using large language models (LLMs)~\cite{zeng2022glm,openai2023gpt4,Claude, geminiteam2024gemini,glm2024chatglm} and large multimodal models (LMMs)~\cite{openai2023gpt4, Claude, hong2023cogagent} as the backbone for these agents~\cite{deng2023mind2web,rawles2023aitw,zhou2023webarena}. 

Despite significant advancements, both training and evaluating mobile agents face challenges, with lacking systematic exploration.  
Previous benchmarks~\cite{rawles2023aitw,sun2022metagui,li-etal-2020-mapping-pixelhelp} often rely on reproducible but static environments, where agents are expected to predict actions based on screenshots without actual interaction. AndroidEnv~\cite{toyama2021androidenv} introduced the first interactive environment for mobile agents and later efforts~\cite{lee2024benchmarking-bmoca,rawles2024androidworlddynamicbenchmarkingenvironment} improved reproducibility but still faced limitations. Moreover, these benchmarks lack systematic evaluation, primarily because almost all recent benchmarks~\cite{yang2023appagent,xing2024understanding,lee2024benchmarking-bmoca,rawles2024androidworlddynamicbenchmarkingenvironment} only tested and implemented prompt-based improvement on closed-source models. This limitation restricts the ability to analyze model behavior, integrate insights, and conduct reinforcement learning experiments effectively. The absence of a unified benchmark comparing open-source and closed-source models across various modalities further exacerbates this issue, limiting opportunities for enhancing open-source solutions.

\begin{figure*}[t]
    \centering
    \begin{subfigure}[t]{.73\textwidth}
        \centering
        \includegraphics[width=\linewidth]{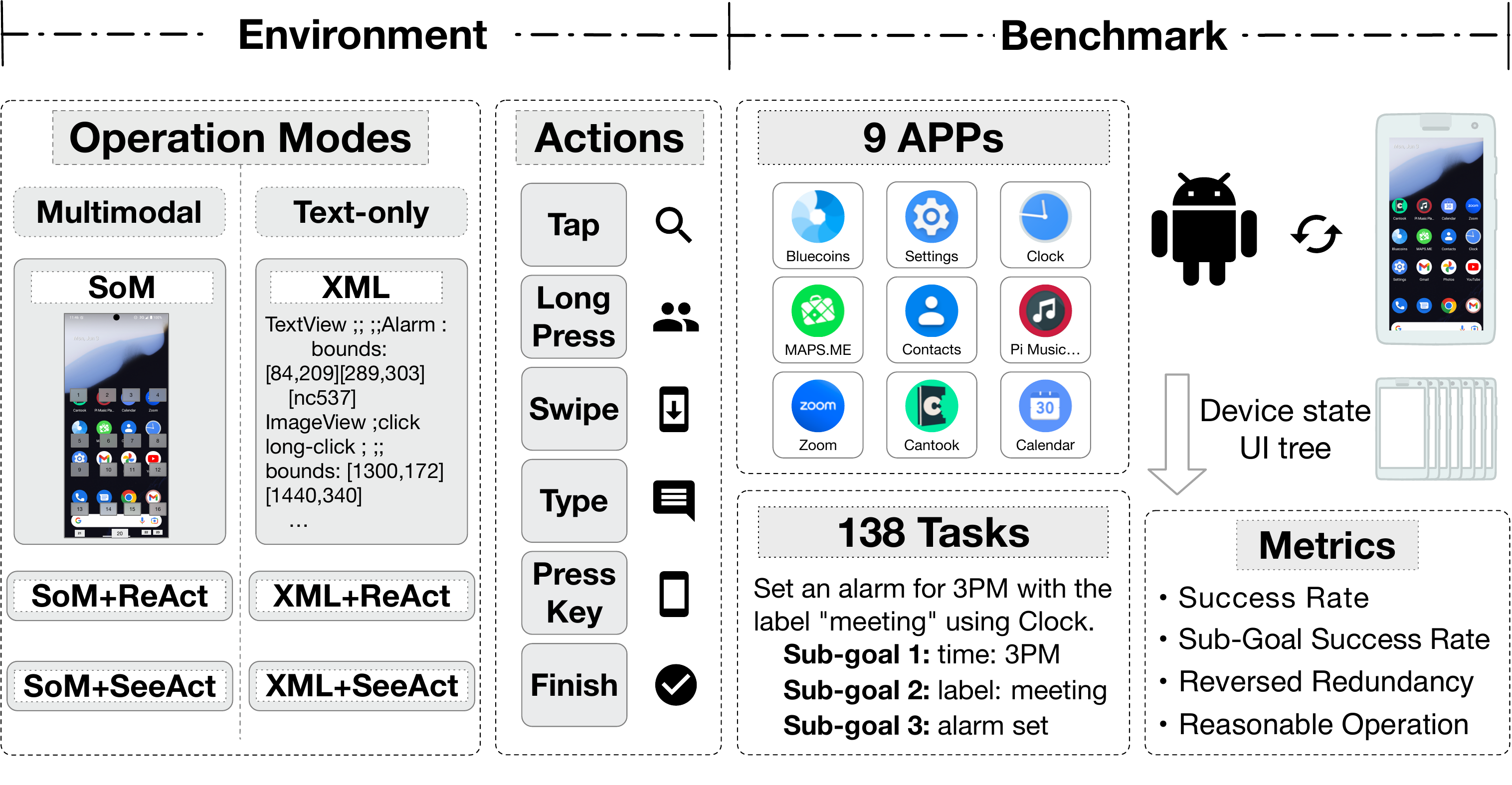}
        \caption{Overview of the environment and benchmark of \textsc{AndroidLab}.}
    \end{subfigure}
    \begin{subfigure}[t]{.26\textwidth}
        \centering
        \includegraphics[width=\linewidth]{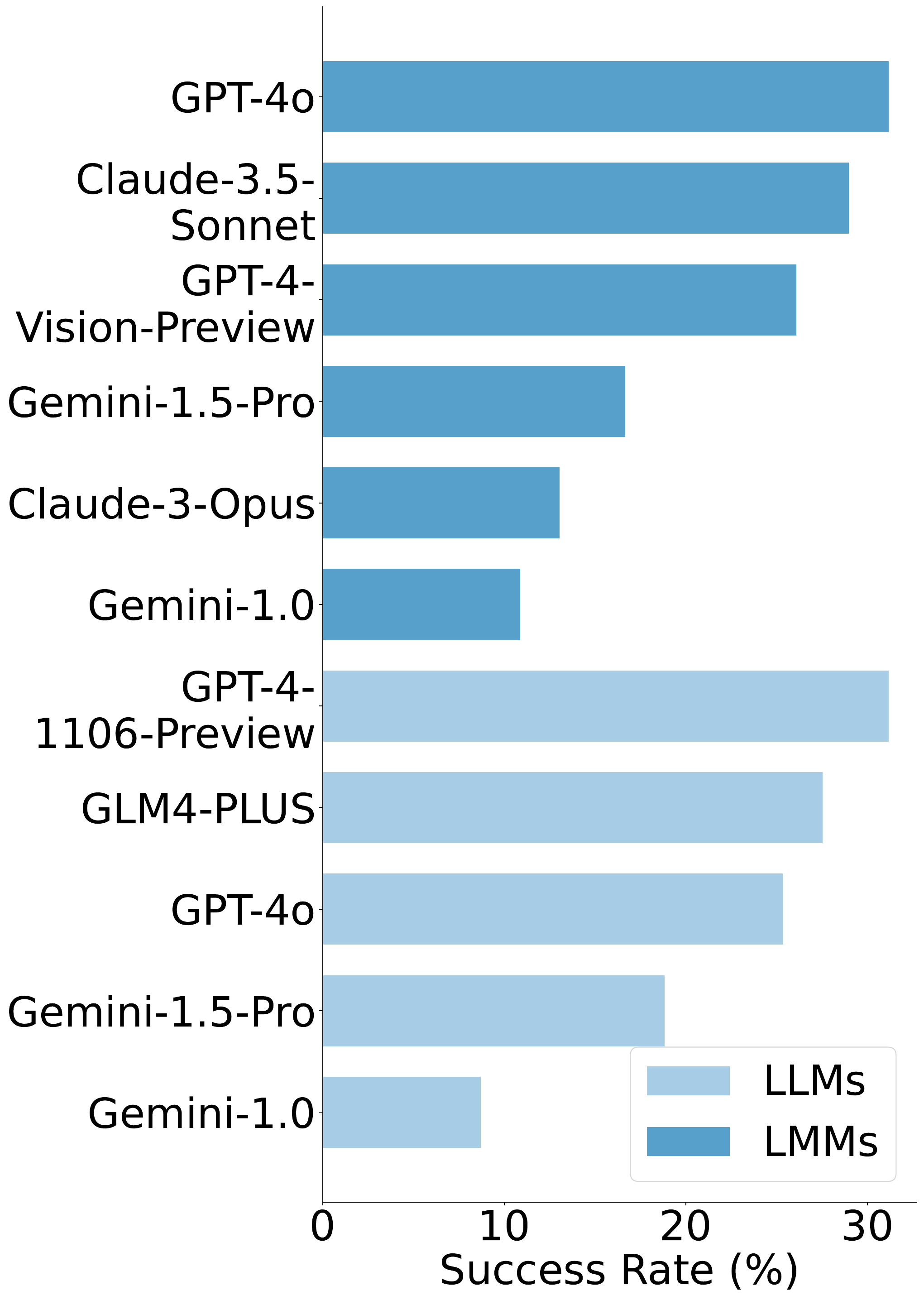}
        \caption{Results of Closed Models.}
    \end{subfigure}
    
    \caption{
    (a) We design the SoM mode for the multimodal models (LMMs) and the XML mode for the text-only models (LLMs), ensuring an identical action space. We also implement ReAct and SeeAct frameworks in both modes. Based on the environment, we propose the \textsc{AndroidLab} benchmark.  (b) \textsc{AndroidLab} benchmark success rates of closed-source models. In the XML mode, GPT-4-1106-Preview has the highest success rate at 31.16\%, the same as GPT-4o in the SoM mode.
    }
    
    \label{fig:main picture}
\end{figure*}

These issues have motivated us to develop a new Android agent evaluation and training framework. In this paper, we propose \textbf{\textsc{AndroidLab}}, which includes a standard operational environment and a benchmark for agents interacting with Android devices. 
We define basic operation modes across LLMs and LMMs by aligning actions and objects within different observations of the mobile system: XML and screenshots, termed XML mode and SoM mode, respectively. Additionally, we introduce two modes for each basic mode, ReAct~\cite{yao2022react} and SeeAct~\cite{zheng2024gpt-seeact}. Node information is annotated in the XML for screenshots using set-of-mark~\cite{yang2023setofmark}, ensuring identical actions across modes for a fair comparison. 
Based on the environment, the \textsc{AndroidLab} benchmark includes 138 tasks across 9 different apps. By utilizing Android virtual devices with preloaded app operation histories and offline data, \textsc{AndroidLab} ensures reproducibility and eliminates external network or time dependencies.

Previous benchmarks had shortcomings in their evaluation metrics, typically provided standardized sequences of operations~\cite{xing2024understanding} or device states~\cite{lee2024benchmarking-bmoca,rawles2024androidworlddynamicbenchmarkingenvironment} as evaluation metrics, which can restrict the diversity of task paths and limit task types to those represented by specific device states. In \textsc{AndroidLab}, each task is divided into multiple required page states as sub-goals, with UI tree structure matching verifying correct traversal. This enables precise assessment of task completion and progress and allows evaluation of nearly all tasks without being constrained by the limitations of system state representations. We also introduce metrics such as reversed redundancy and reasonable operation to evaluate action efficiency.

We have evaluated 17 open-source and closed-source models using the \textsc{AndroidLab} benchmark. Although the GPT series achieved over 30\% success rate in both XML and SoM modes, we observed that open-source models performed poorly, with the best reaching only around 5\% success rate. Initial attempts to enhance mobile agent performance through more complex reasoning frameworks led to marginal improvements despite significantly increased inference times. Therefore, fine-tuning small-scale open-source models may bridge the gap to closed-source performance, enhancing mobile agent accessibility.

\begin{figure*}[t]
    \centering
    \begin{subfigure}[t]{.42\textwidth}
        \centering
        \includegraphics[width=\linewidth]{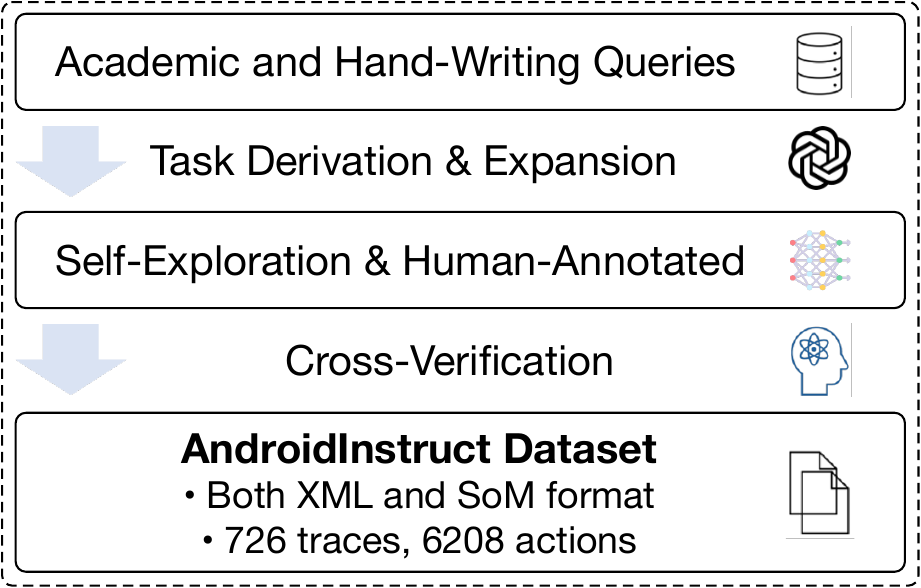}
        \caption{Overview of Android Instruct data collection.}
    \end{subfigure}
    \begin{subfigure}[t]{.57\textwidth}
        \centering
        \includegraphics[width=\linewidth]{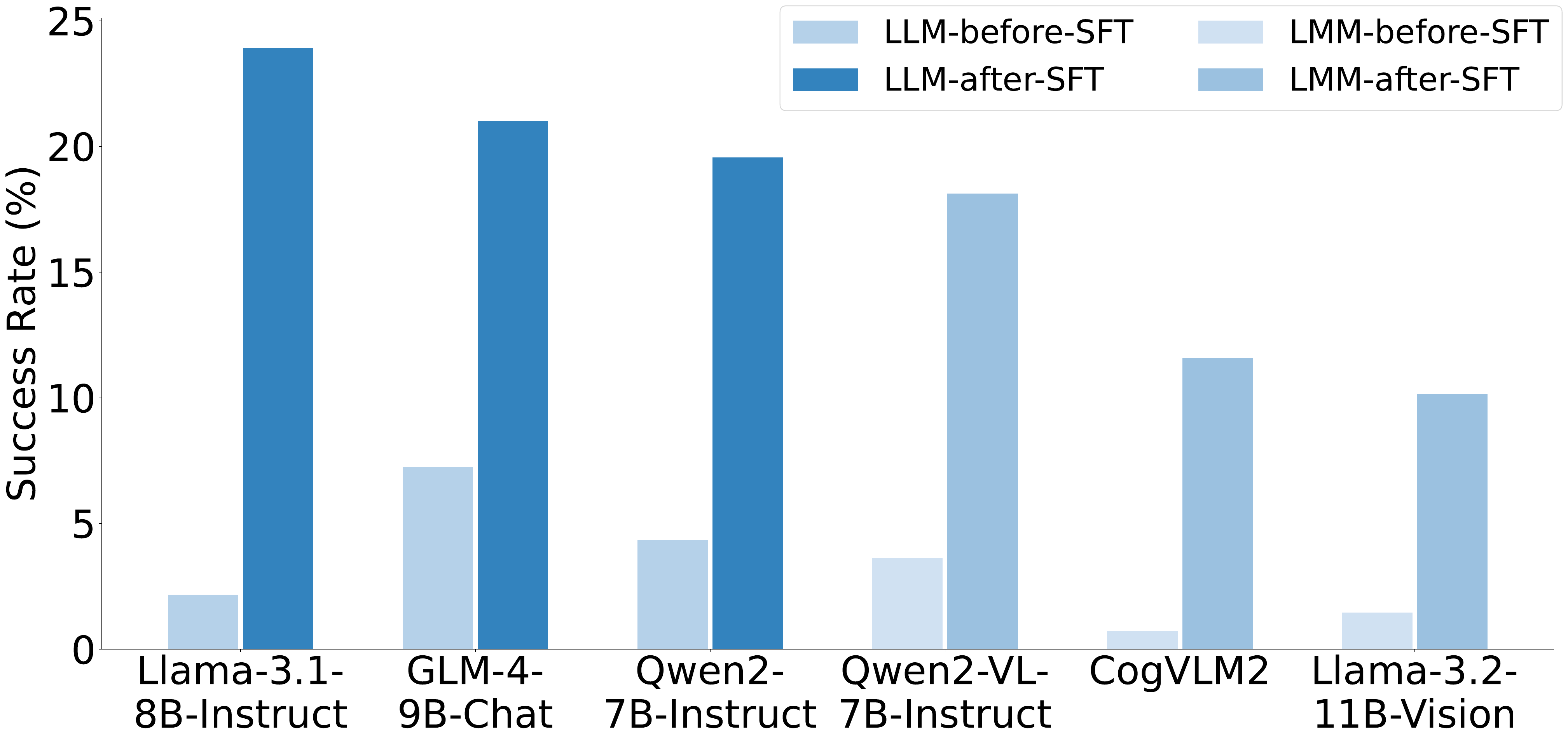}
        \caption{Success Rates of before and after fine-tuning by Android Instruct.}
    \end{subfigure}
    
    \caption{
    (a) We have collected over 726 traces containing more than 6208 fully aligned steps of XML and SoM mode training data.  (b) By using the Android Instruct dataset, we trained six open-source text-only and multimodal models, achieving an average success rate from 4.59\% to 21.50\% for LLMs and from 1.93\% to 13.28\% for LMMs. respectively, reaching a performance level comparable to proprietary models.
    }
    
    \label{fig: before after sft}
\end{figure*}
By using \textsc{AndroidLab}'s operation modes and action space, we have constructed the Android Instruct dataset. We develop an online annotation tool with the same action space, collecting 10.5k traces and 94.3k steps from annotators. Among these, 6208 steps are derived from the Apps included in the \textsc{AndroidLab} benchmark, and we use this portion of the data to fine-tune the model. This dataset includes tasks, phone screen states, XML information, and operations, and has been used to fine-tune six text-only and multimodal models. As shown in Figure~\ref{fig: before after sft}, fine-tuning with our dataset raises average success rates from 4.59\% to 21.50\% for LLMs and from 1.93\% to 13.28\% for LMMs. Our further analysis reveals that fine-tuning improves operational accuracy, efficiency, and reduces redundancy in Android agents.

The contributions are summarized as follows:
                                            
\begin{itemize}
[leftmargin=1.5em,itemsep=0pt,parsep=0.2em,topsep=0.1em,partopsep=0.0em]
\item[$\bullet$] We design the \textsc{AndroidLab} suite, which includes a standard operational environment and a benchmark. This suite unifies the evaluation and training of Android Agents, as shown in Figure~\ref{fig:main picture}.

\item[$\bullet$] We develop \textsc{AndroidLab} benchmark, a reproducible and challenging benchmark for evaluating mobile agent capabilities. It includes a simulated evaluation environment and 138 tasks, as shown in Figure~\ref{fig:apps and subcategories} based on text-only or multimodal inputs. \textsc{AndroidLab} benchmark presents significant challenges, as the leading model GPT-4o only achieves 31.16\%. Parts of the AndroidLab benchmark's SoM modes are also included in the VisualAgentBench~\cite{liu2024visualagentbench} as the VAB-Mobile component.

\item[$\bullet$] We construct an Android Instruct dataset, containing 94.3k operation records for fine-tuning. This dataset supports both text-only and multimodal training, yielding competitive results in LLM and LMM models, as shown in Table \ref{table:main-result}. We also demonstrate that fine-tuned models achieve comparable scores and offer the best balance of efficiency and accuracy.

\begin{figure*}[t]
    \centering
    \includegraphics[width=\textwidth]{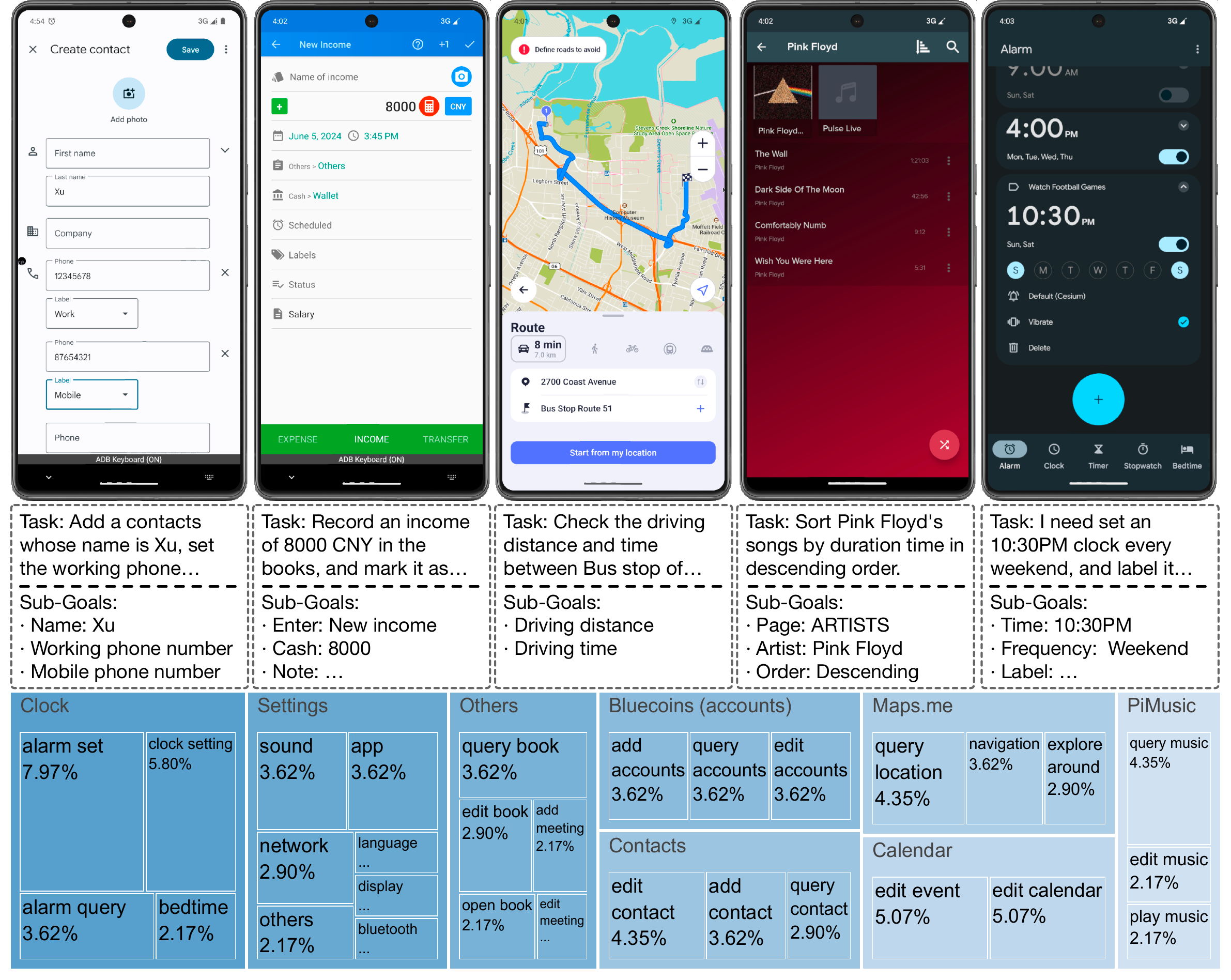}
    \caption{Task examples and the distribution of all apps and subcategories in the \textsc{AndroidLab} benchmark. We decomposed each task into sub-goals and evaluated them independently. A task is considered complete only if all sub-goals are correctly addressed.
}
    \label{fig:apps and subcategories}
\vspace{-2mm}
\end{figure*}
\
\end{itemize}

\section{Retated Work}
\vpara{Benchmarks for Agents.}
Recent advancements in large foundation models have led to new agent benchmarks tailored to these models. Agents interact with external environments primarily through writing code~\cite{chen2021humaneval,zheng2023codegeex,zhang2024naturalcodebench,austin2021program-mbpp} or invoking APIs~\cite{guo2024stabletoolbench,li2023apibank,peng2021revisiting-apibench}. Specialized benchmarks have been designed for interaction with operating systems, categorized into Desktop and Mobile. For Desktop, static benchmarks~\cite{mialon2023gaia,deng2023mind2web,kapoor2024omniact} evaluate agents by single-step operation or operations sequence without a virtual environment. Otherwise, dynamic benchmarks provide interactive web browser~\cite{liu2018reinforcement-miniwob,zhou2023webarena,yao2022webshop,koh2024visualwebarena} or Unix-like system virtual environment~\cite{hong2023cogagent,OSWorld}, making evaluation more flexible and realistic.

Mobile benchmarks for Android began with static systems like PixelHelp~\cite{li-etal-2020-mapping-pixelhelp} and MetaGUI~\cite{sun2022metagui} and later expanded through AITW~\cite{rawles2023aitw}, which provided over 5 million images. AndroidEnv~\cite{toyama2021androidenv} introduced dynamic evaluations, while Android Arena~\cite{xing2024understanding} added cross-app evaluations. Although task diversity was limited, B-MOCA~\cite {lee2024benchmarking-bmoca} standardized the Android Virtual Device. AndroidWorld~\cite{rawles2024androidworlddynamicbenchmarkingenvironment} offers reward signals for 116 tasks across 20 real-world apps but does not support instruction-tuning data construction.

\vpara{Agents for Interactive System.} For Web environments, WebGPT~\cite{nakano2021webgpt} and WebGLM~\cite{liu2023webglm} integrate LLMs for improved question-answering. MindAct~\cite{deng2023mind2web}, WebAgent~\cite{gur2023real}, and AutoWebGLM~\cite{lai2024autowebglm} focus on executing complex interactive tasks. In mobile agents, early work on Android systems utilized multiple execution modules~\cite{burns2021mobile-motif,venkatesh2023ugif,li-etal-2020-mapping-pixelhelp,zhan2023autoui}. PixelHelp~\cite{li-etal-2020-mapping-pixelhelp} mapped actions to images, while Auto-GUI~\cite{zhan2023autoui} used image and text encoders with LLMs for CoT outputs. CogAgent~\cite{hong2023cogagent} achieved SOTA on AITW~\cite{rawles2023aitw} by combining modules for action prediction. Recent zero-shot mobile agents using GPT-4V~\cite{openai2023gpt4} have shown strong results~\cite{yang2023appagent,zheng2024gpt-seeact,yan2023gpt4v-gui,wang2023enabling}, but planning complexity limits inference speed and practical deployability due to security restrictions.

\section{\textsc{AndroidLab}}

\subsection{The Operation Environment}

\textsc{AndroidLab} defines a set of action spaces and two operation modes, forming the \textsc{AndroidLab} environment. We adopt the main action space from prior work and add a model return value (finish action). The two basic operation modes are SoM~\cite{yang2023setofmark} and XML-only, differing in whether the agent can access a snapshot of the phone screen. For comparison, we also implement ReAct~\cite{yao2022react} and SeeAct~\cite{zheng2024gpt-seeact}. This framework supports real and virtual Android devices and is compatible with Android-like mobile operating systems.

\vpara{Action Space.} Based on the action spaces from AppAgent~\cite{yang2023appagent} and Android Env~\cite{toyama2021androidenv}, we define four basic phone operations: \texttt{Tap}, \texttt{Swipe}, \texttt{Type}, \texttt{Long Press}, along with two shortcut keys, \texttt{Home} and \texttt{Back}, as the core action space. We add the \texttt{Finish} action as the final step, allowing the agent to return execution results or answers. This action space applies to all modes.

\vpara{XML Mode.} XML mode is tailored for text-only input models (LLM). Inspired by Android Arena~\cite{xing2024understanding}, we redesign the XML compression algorithm to convey screen information. The LLM selects corresponding elements directly for operations.

\vpara{SoM Mode.} SoM mode is for multimodal input models (LMM), based on the Set-of-Mark method~\cite{yang2023setofmark}. Each clickable or focusable element is assigned a serial number, and the LMM selects the element by its number. The selected elements in SoM mode align with those in the compressed XML list, allowing both modes to interact with the same action space and objects.

These basic operation modes directly require the agent to output operation commands. Based on these two methods, we further test two novel agent frameworks, ReAct~\cite{yao2022react} and SeeAct~\cite{zheng2024gpt-seeact}. These two frameworks allow the agent to observe and reflect on the environment or more easily select specific tasks to execute. Please refer to Appendix~\ref{appendix:mode} for more details about our operation modes.

\vpara{ReAct modes.} Based on the above two modes, we follow~\cite{yao2022react} to prompt the model, allowing models to think step by step and output their thought and reasoning process. 

\vpara{SeeAct modes.} Following~\cite{zheng2024gpt-seeact}, we separate the reasoning and element grounding process. We instruct models to interact for two rounds in a single operation. The models are supposed to generate a detailed description of the desired action and output the real action, respectively.

\subsection{The Reproducible Benchmark}

Based on the environment, \textsc{AndroidLab} benchmark offers a deterministic and reproducible evaluation platform, allowing users to perform fair and challenging comparisons of Android agent capabilities. \textsc{AndroidLab} benchmark introduces the following designs:

\begin{itemize}[leftmargin=*,itemsep=0pt,parsep=0.2em,topsep=0.2em,partopsep=0.0em]
    \item We gathered 138 tasks from nine apps, ensuring reproducibility. These tasks, derived from common mobile scenarios, are divided into two types: (a) Operation Tasks, where agents must complete a series of actions to meet a goal, and (b) Query Tasks, where agents answer queries based on phone information.
    \item Using phone XML data, we identify screen information that uniquely defines task completion, making task completion our primary metric. Additionally, we select auxiliary metrics such as the proportion of valid actions and the redundancy of successful operation sequences.
\end{itemize}

\subsubsection{Task Formulation}

We formalize each task input as a $4$-tuple: $\text{Task}(E, I, F, M)$. Here, $E$ represents the execution environment of the task, which, in the context of benchmark testing, is the pre-packaged AVD (Android virtual device) image. This includes a fixed phone screen size, Android version, API level, and a fixed app usage state. $I$ denotes the specific natural language instruction for the task. To avoid confusion during testing, we specify the app required to complete the task in natural language. $F$ represents the agent testing framework. Finally, $M$ denotes the backbone model used to perform the task, primarily referring to LLMs or LMMs.

Thus, we can formally define the two types of tasks included in \textsc{AndroidLab}:

\vpara{Operation Task.}
$\text{T}(E, I, F, M) \to (S_1, \ldots, S_n)$. The output of this type of task is a sequence of continuous Android virtual machine states.

\vpara{Query Task.}
$\text{T}(E, I, F, M) \to (S_1, \ldots, S_n, A)$. This type of task assesses the agent's ability to answer specific questions based on the state sequence after exploration. The model must explore the environment to find the answers and output the correct response.

Based on the above formulation, we design 138 tasks, including 93 Operation Tasks and 45 Query Tasks. Please refer to Appendix~\ref{appendix:all task} for detailed information.

\subsubsection{Reproducible Designs}

To ensure our evaluation reflects real-world agent usage scenarios with an appropriate level of difficulty and full reproducibility, we design the tasks with the following considerations:
\begin{itemize}[leftmargin=*,itemsep=0pt,parsep=0.2em,topsep=0.2em,partopsep=0.0em]
    \item \textbf{Fixed Evaluation Time and Space}: We use ADB commands at the start of each evaluation to set the machine's time and virtual geolocation to predetermined values.

    \item \textbf{Offline Testing}: All test apps function offline, with preloaded usage records in the AVD image to ensure normal usability without an internet connection.

    \item \textbf{Predefined Answers}: For query-based tasks, we conduct operations on the corresponding apps in advance to guarantee uniquely determined correct results.
\end{itemize}

\subsubsection{Metrics}

Previous evaluations with virtual environments have relied on indirect metrics like single-step accuracy and operation path matching, leading to imprecise assessments. In response, \textsc{AndroidLab} benchmark introduces a task-completion-based evaluation system that judges directly from device and screen states. Our key metrics are:

\begin{figure*}[t]
    \centering
    \includegraphics[width=\textwidth]{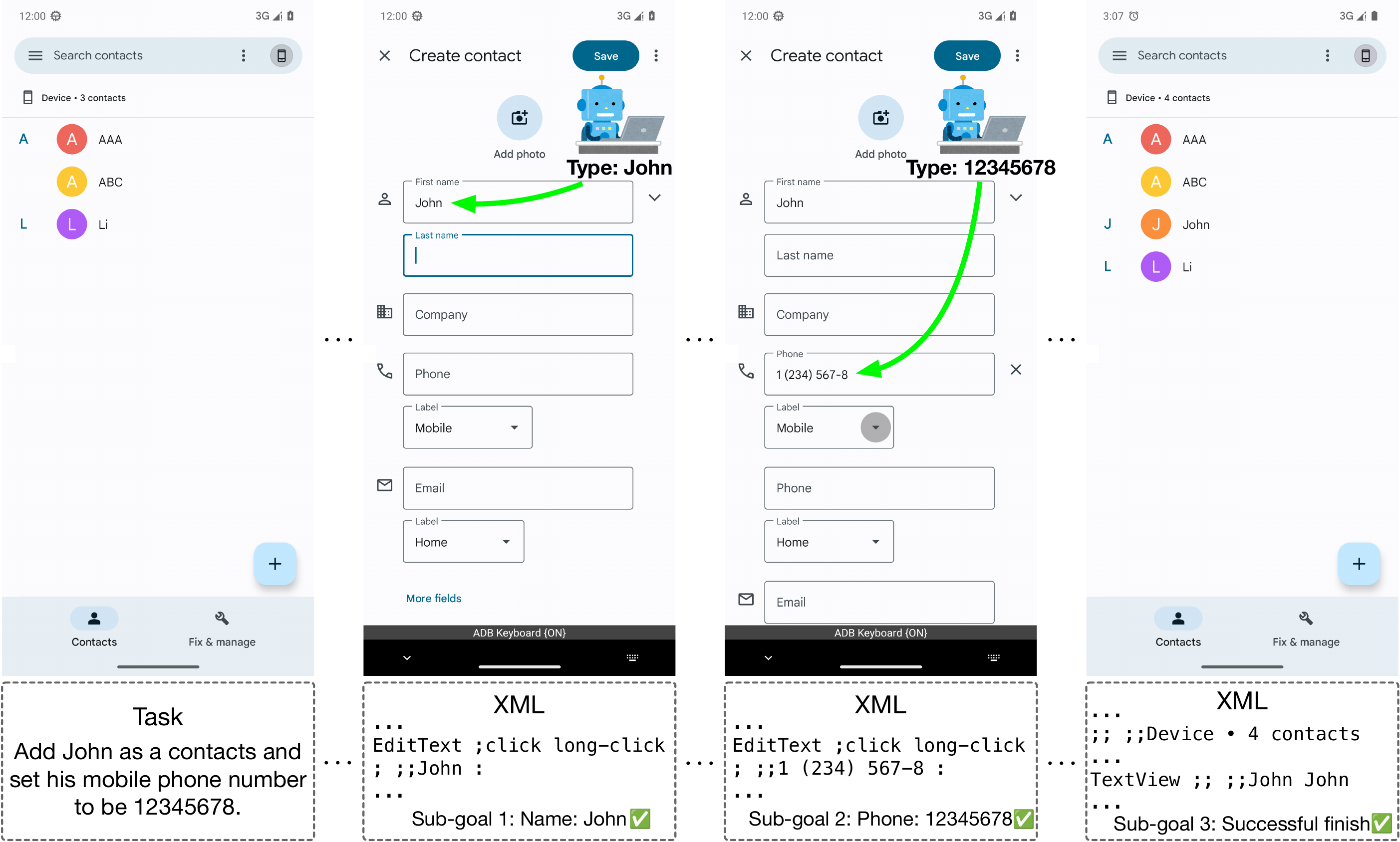}
    \caption{An example of an agent completing all sub-goals of the entire task. We only present the starting and ending steps, along with the steps where the agent completes each sub-goal. It is essential that we record the completion status of each sub-goal. Without this information, we may not be able to obtain detailed information from the XML of the finished page, which could lead to a misjudgment of the task.}
    \label{fig:sub-goals}
\vspace{-2mm}
\end{figure*}


\begin{itemize}[leftmargin=*,itemsep=0pt,parsep=0.2em,topsep=0.2em,partopsep=0.0em]
    \item \textbf{Success Rate}: For Operation Tasks, we divided a complete task into multiple sub-goals and identified the specific page information for each sub-goal completion. By checking and matching specific UI tree elements, we assess each sub-goal completion status individually. The task is considered successfully executed when all sub-goals are completed. We have also set up a few tasks that can directly use the device state to determine if they were completed correctly. For Query Tasks, advanced LLMs verify if the model’s predicted results match the standard answers, avoiding errors from direct string comparisons. We provide an example in Fig~\ref{fig:sub-goals}.

    \item \textbf{Sub-Goal Success Rate}: Tasks are decomposed into sub-goals, and completion is assessed sequentially. This finer metric rewards models with stronger understanding and operational capabilities. Only Operation Tasks include the Sub-Goal Success Rate.

    \item \textbf{Reversed Redundancy Ratio}: As in prior work~\cite{xing2024understanding}, redundancy is measured by comparing the model’s operation path length to a human benchmark. We calculate this for completed tasks and take the reciprocal, so higher values indicate less redundancy. We do not report SR < 5 because there are too few completed tasks, which may be affected by a small number of special values. It should also be emphasized that this metric may exceed 100 because the steps of human operation are not necessarily optimal.

    \item \textbf{Reasonable Operation Ratio}: This metric evaluates the proportion of operations after which the screen changed. Unchanged screens indicate the operation was ineffective and thus deemed unreasonable.
\end{itemize}

By incorporating these metrics, our evaluation system provides a comprehensive and precise assessment of an agent's performance in completing specified tasks.

\section{Android Instruction Data}

\begin{figure*}[htbp]
    \centering
    \begin{subfigure}[b]{0.32\textwidth}
        \includegraphics[width=\textwidth]{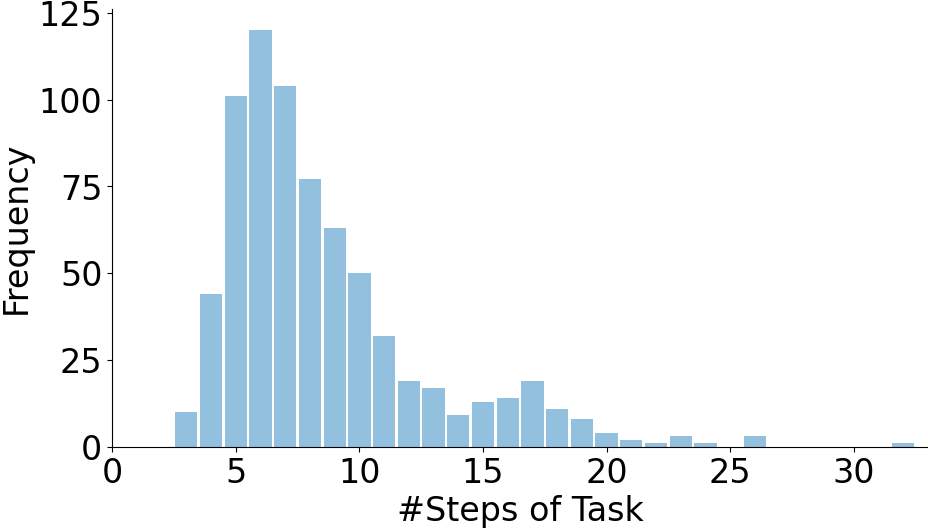}
        \caption{Step Distribution Across Tasks}
    \end{subfigure}
    \hfill
    \begin{subfigure}[b]{0.32\textwidth}
        \includegraphics[width=\textwidth]{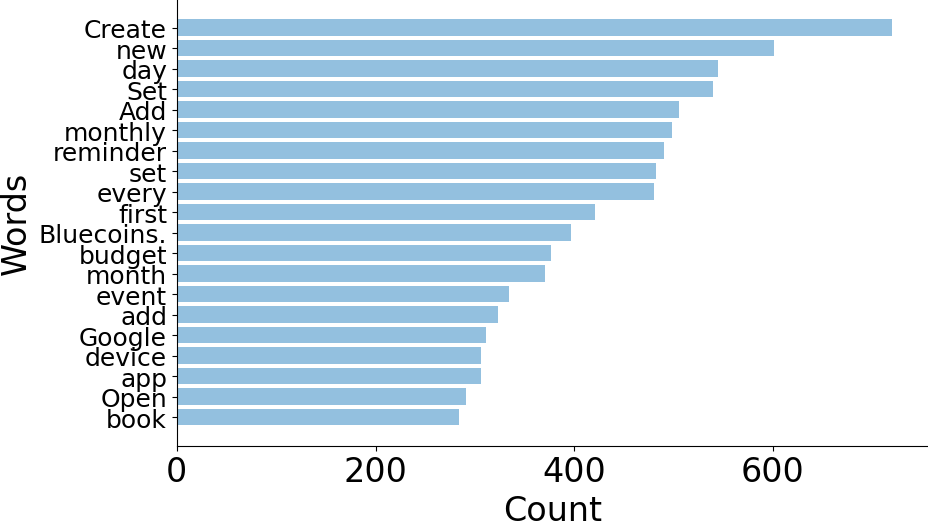}
        \caption{Top 20 Words in Instructions.}
    \end{subfigure}
    \hfill
    \begin{subfigure}[b]{0.32\textwidth}
        \includegraphics[width=\textwidth]{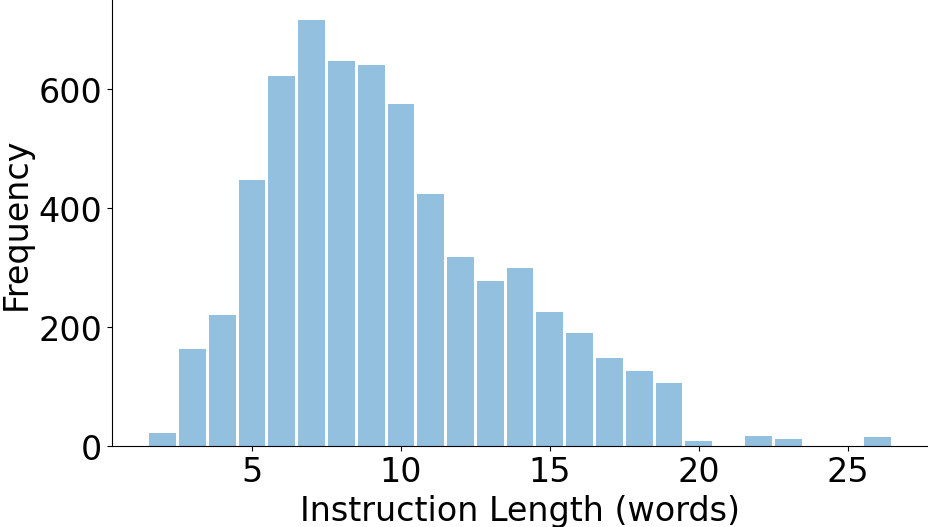}
        \caption{Instruction Length Distribution.}
    \end{subfigure}

    \vspace{1cm} 

    \begin{subfigure}[b]{0.32\textwidth}
        \includegraphics[width=\textwidth]{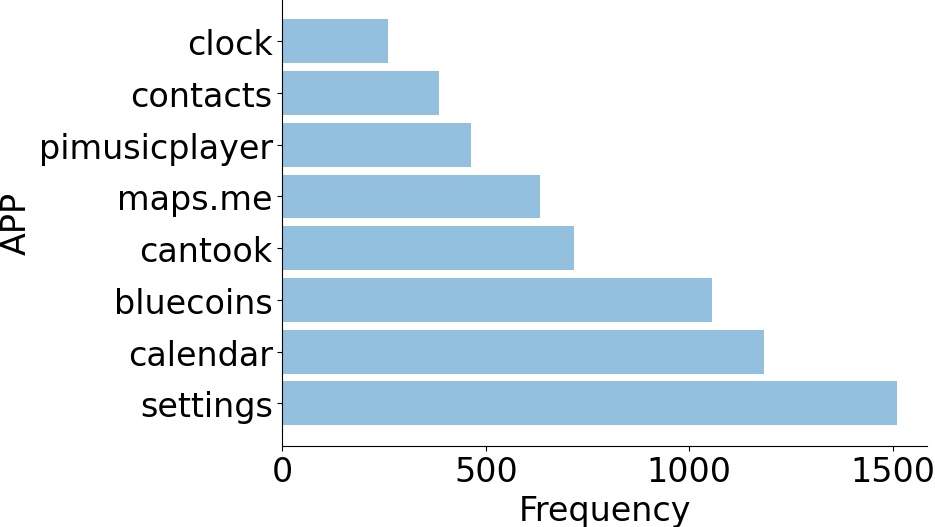}
        \caption{APP Distribution.}
    \end{subfigure}
    \hfill
    \begin{subfigure}[b]{0.32\textwidth}
        \includegraphics[width=\textwidth]{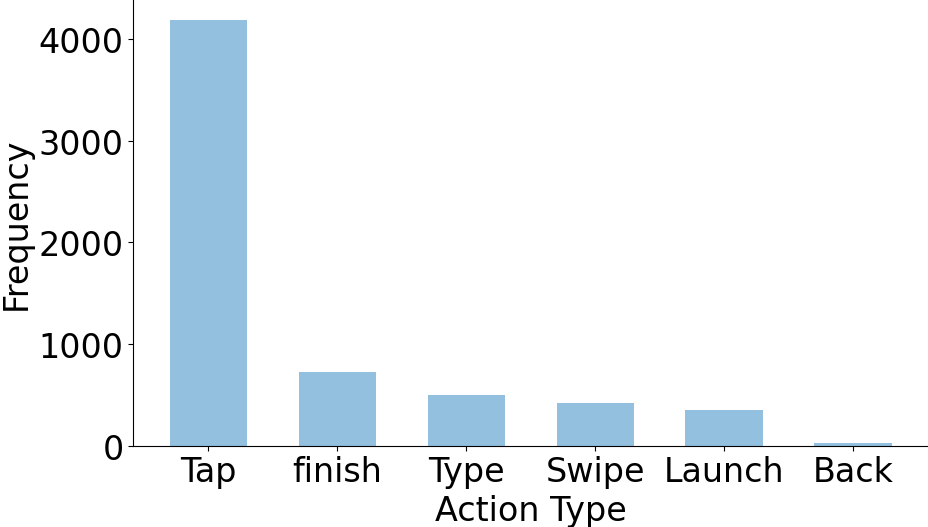}
        \caption{Actions Distribution.}
    \end{subfigure}
    \hfill
    \begin{subfigure}[b]{0.32\textwidth}
        \includegraphics[width=\textwidth]{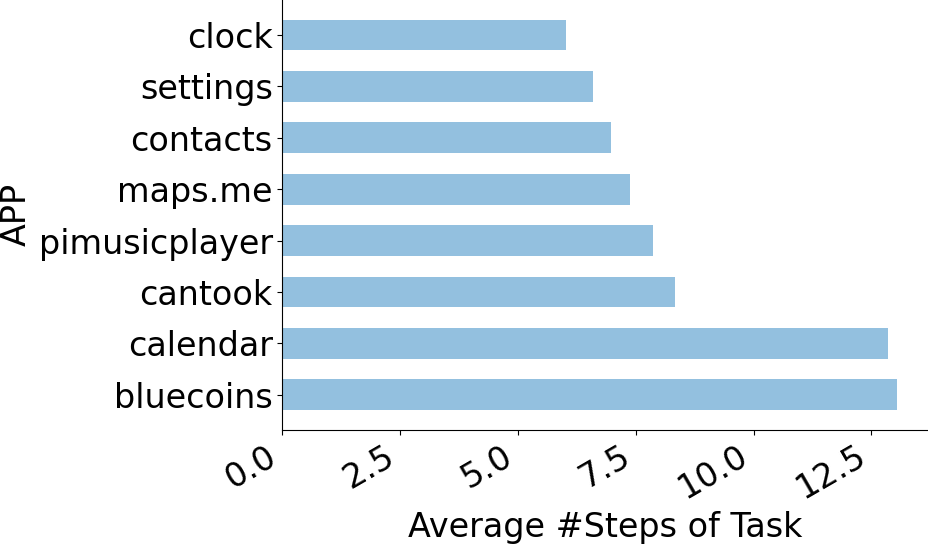}
        \caption{Average Task Length per App}
    \end{subfigure}

    \caption{Statistics for Android Instruct dataset. We collect 726 traces and 6208 steps across Apps in \textsc{AndroidLab} benchmark.}
    \label{fig:statistics_AID}
\end{figure*}

Building an open-source, deployable Android operation agent is a significant challenge in AI research. Previous work on Android agents has focused on using powerful closed-source models to design interaction logic~\cite{zheng2024gpt-seeact,yang2023appagent,wang2023enabling}, raising concerns about accessibility, privacy, and efficiency. To address this, we aim to build an open-source mobile agent. The main challenge lies in generating training data for mobile operations to handle open-world tasks in diverse environments.

We propose task derivation and expansion methods for task generation, allowing models to generate tasks for specific apps controllably. \textsc{AndroidLab} connects to devices via ADB, enabling compatibility with various real or virtual devices for data generation. Using self-exploration and manual annotation, we generate example operation traces. To make it easier for annotators to work on real devices (rather than emulators), we developed an online annotation tool. This tool uses ADB commands to monitor user interactions on the phone and captures screenshots and page XML before each action. Our Android Instruction data is built on the Task (E, I, F) framework within \textsc{AndroidLab}'s environment.

\subsection{Data Construction}

The primary challenges in data construction include generating executable Android instructions and annotating operation path data. Our approach involves three steps:

1. \textbf{Task Derivation and Expansion}: We use academic datasets~\cite{rawles2023aitw,coucke2018snips} and manually write instructions to seed task generation. Language models are employed to create additional tasks, which are reviewed and added to the dataset, ensuring realistic and executable instructions.

2. \textbf{Self-Exploration}: LLMs and LMMs are used for automatic task exploration, outputting \textit{finish} when done. Initially, manual selection was used to verify results, but a reward model later replaced it after gathering 500 traces.

3. \textbf{Manual Annotation}: This process involves four steps: (1) \textbf{Instruction Check}, where annotators evaluate the feasibility of the given task; (2) \textbf{Preliminary Familiarization}, allowing them to explore the app interface before performing tasks; (3) \textbf{Task Execution}, in which the annotators execute and document each task step; and (4) \textbf{Cross-Verification}, where a second annotator reviews the task trace to ensure its accuracy.

This combination of autonomous and manual processes resulted in 10.5k traces and 94.3k steps, and we use 726 traces and 6208 steps derived from the Apps included in the \textsc{AndroidLab} benchmark for training. We provide statistics of the Android Instruct dataset in Fig~\ref{fig:statistics_AID}. More details are in Appendix~\ref{appendix:training data}.

\subsection{Annotation Tool}

To more accurately and efficiently record operation trajectories and page information (XML), we design an annotation tool.

\vpara{Acquisition of Page Information:} Android Debug Bridge (ADB) is currently the most widely used tool for obtaining page information~\cite{yang2023appagent,rawles2024androidworlddynamicbenchmarkingenvironment}. ADB is a versatile command-line utility that retrieves the XML data of the current page. However, when dealing with a diverse range of mobile applications, ADB sometimes fails to acquire the XML for certain pages. Specifically, ADB waits for all UI components on the page to become idle before retrieving component information. If this process exceeds a predefined time limit, ADB stops the XML acquisition. This issue is particularly evident on mobile pages with dynamic components, such as playback bars and animations in audio players, where continuously active elements prevent ADB from obtaining the XML. To address this, we reimplemented the XML acquisition functionality using the Android Accessibility Service, allowing annotators to determine the appropriate timing for retrieving page XML.

\vpara{Recording Operation Trajectories:} We mainly need to record three types of user actions: clicks, swipes, and text input. For click actions and swipe actions, annotators complete the actions directly on the phone, while we use ADB commands to capture screen events. Based on the press, release positions, and duration of these events, we determine whether the action was a click or swipe. For text input, we utilize the ADB keyboard to complete the entire input in a single operation, minimizing the number of annotations required. Before each action, the user must first use the annotation tool to record the current page information, ensuring that the recorded page data matches the context observed during human interaction.

\subsection{Training}

To explore the effectiveness of our dataset on lightweight open-source models, we select Llama-3.1-8B-Instruct, GLM-4-9B-Chat, Qwen2-7B-Instruct, Llama-3.2-11B-Vision-Instruct, Qwen2-VL-7B-Instruct and CogVLM2 (cogvlm2-llama3-chat-19B) as the training backbones for LLM and LMM, respectively. Due to our preliminary experiments showing that training agents from base models yield better results, we select the base versions of all models for fine-tuning, except for Qwen2-VL-7B-Instruct (as no open-source base model is available). However, we still report the instruct versions as baselines because the base models cannot follow instructions without further tuning. For all training sessions, we use a batch size of 32 and a maximum sequence length of 4096, training for five epochs. The learning rate is set to 1e-5.

\begin{table*}[h]
    \centering
    \setlength{\tabcolsep}{12.5pt}

    \caption{\textbf{Main Result of XML and SoM modes.} SR, Sub-SR, RRR, and ROR stand for Success Rate, Sub-Goal Success Rate, Reversed Redundancy Ratio, and Reasonable Operation Ratio, respectively. For all these metrics, a higher value means better. \textbf{-ft} represents a finetuned model. In each mode, \textbf{Bold} represents the best result. We do not report RRR score if SR < 5.}
    \label{table:main-result}
   
    \begin{tabular}{@{}clcccc@{}}
    \toprule
    \textbf{Mode} & \multicolumn{1}{c}{\textbf{Model}} & \textbf{SR}          & \textbf{Sub-SR}      & \textbf{RRR}          & \textbf{ROR}         \\ \midrule
    \multirow{8}{*}{XML}    & GPT-4o                                 & 25.36                 & 30.56                & \textbf{107.45}       & 86.56                \\
                            & GPT-4-1106-Preview                     & \textbf{31.16}        & \textbf{38.21}        & 66.34                 & 86.24                \\
                            & Gemini-1.5-Pro                   & 18.84                 & 22.40                 & 57.72                 & 83.99                \\
                            & Gemini-1.0                       & 8.70                  & 10.75                 & 51.80                 & 71.08   \\  
                            & GLM4-PLUS                                      & 27.54                 & 32.08                 & 92.35                 & 83.41 \\
                            & LLaMA3.1-8B-Instruct                   & 2.17                  & 3.62                  & -                & 52.77               \\
                            & Qwen2-7B-Instruct                   & 4.35                  & 4.95                 & -               & 67.26               \\
                            & GLM4-9B-Chat & 7.25                  & 9.06                  & 54.43                 & 58.34                \\ \cdashline{2-6} \\[-0.75em]

  \multirow{3}{*}{XML+SFT}  & LLaMA3.1-8B-\textbf{ft}     & 23.91                  & 30.31                 & 75.58                 & 92.46           \\
  & Qwen2-7B-\textbf{ft}  & 19.57               & 24.40             & 77.31                & 92.48 \\
  & GLM4-9B-\textbf{ft} & 21.01                  & 26.45                 & 74.81                &    \textbf{93.25}            \\ \midrule

    \multirow{9}{*}{SoM}    & GPT-4o                                 & \textbf{31.16}        & \textbf{35.02}        & 87.32                 & 85.36                \\
                            & GPT-4-Vision-Preview                   & 26.09 & 29.53 & 99.22 & 78.79                \\
                            & Gemini-1.5-Pro                   & 16.67                 & 18.48                 & 105.95       & \textbf{91.52}       \\
                            & Gemini-1.0                       & 10.87                 & 12.56                 & 72.52                 & 76.70                \\
                            & Claude-3.5-Sonnet                              & 28.99                  & 32.66                   & \textbf{113.41}                & 81.16                \\
                            & Claude-3-Opus                                 & 13.04                 & 15.10                 & 81.41                 & 83.89                \\
                            & CogVLM2                                  & 0.72                 & 0.72                 & -                 & 17.97                \\
                            & LLaMA3.2-11B-Vision-Instruct                                   & 1.45                 & 1.45                 & -                 & 50.76                \\
                            & Qwen2-VL-7B-Instruct    & 3.62                 & 4.59                 & -                 & 84.81                \\
                            \cdashline{2-6} \\[-0.75em]
                           
    \multirow{3}{*}{SoM+SFT}                        & CogVLM2-\textbf{ft}                  & 11.59                 & 16.06                 & 57.37                 & 85.58  \\   
    & LLaMA3.2-11B-Vision-\textbf{ft}                  & 10.14                 & 12.98                & 61.67                 & 87.85  \\   
    & Qwen2-VL-7B-Instruct-\textbf{ft}                  & 18.12                & 22.64                & 65.23                & 88.29 
                   \\ \bottomrule
    \end{tabular}
\end{table*}

\section{Experiments}

\subsection{Experiment Setup}

\vpara{Evaluation Settings.} 
In preliminary tests, we found that even though we specified the use of certain apps in the instructions, agents failed to complete tasks because they could not launch the respective apps correctly. To avoid errors caused by a single reason, we start tasks directly within the specified app in the formal experiments and then allow the agent to proceed. Additionally, we set a maximum execution step limit of 25 for each task, with a 3-second interval for the virtual machine to respond to each operation. We generate by greedy search for each task of all models.

\vpara{Baseline Models.} 
For large language models (LLMs) with text-only input capability, we selected GPT-4o~\cite{openai2023gpt4}, GPT-4-1106-Preview~\cite{openai2023gpt4}, Gemini-1.5-Pro~\cite{geminiteam2024gemini}, Gemini-1.0~\cite{geminiteam2024gemini}, GLM-4-PLUS~\cite{glm2024chatglm},  Llama-3.1-8B-Instruct ~\cite{touvron2023llama}, GLM-4-9B-Chat~\cite{glm2024chatglm} and Qwen2-7B-Instruct~\cite{bai2023qwen} as the baselines for testing in the XML mode. For large multimodal models (LMMs) with image input capability, we chose GPT-4o~\cite{openai2023gpt4}, GPT-4-Vision-Preview~\cite{openai2023gpt4}, Gemini-1.5-Pro~\cite{geminiteam2024gemini}, Gemini-1.0~\cite{geminiteam2024gemini}, Claude-3.5-Sonnet, Claude-3-Opus~\cite{Claude}, Llama-3.2-11B-Vision-Instruct~\cite{touvron2023llama}, Qwen2-VL-7B-Instruct~\cite{wang2024qwen2} and CogVLM2~\cite{wang2023cogvlm} as the baselines for testing in the SoM mode. We also further evaluated the performance of GPT-4o and Gemini-1.5-Pro under the ReAct and SeeAct frameworks in both modes.

\subsection{Main Results}

As shown in Table~\ref{table:main-result}, in the XML mode, GPT-4-1106-Preview outperforms the other models with a Success Rate (SR) of 31.16\%, the highest in this mode while also achieving the best Sub-Goal Success Rate (Sub-SR) at 38.21\%. Although GPT-4o exhibits slightly lower SR (25.36\%), it achieves the highest Reversed Redundancy Ratio (RRR) at 107.45, indicating its strong ability to reduce unnecessary operations. The ROR metric shows that both models in the GPT-4 series perform comparably, with around 86\% of operations being reasonable but with room for improvement in efficiency. Other models, such as Gemini-1.5-Pro, show moderate performance, with ROR around 80, but lag in SR.

In the SoM mode, GPT-4o again shows dominance, reaching an SR of 31.16\% and a Sub-SR of 35.02\%, the highest in both categories. GPT-4-Vision-Preview follows closely, but models like Claude-3.5-Sonnet exceeded GPT-4o in RRR (113.40), demonstrating a higher efficiency in task completion with fewer redundant steps. The Reasonable Operation Ratio in SoM mode indicates that models such as tuned LLaMA3.2-11B-Vision achieve the best ROR at 92.57\%, showing the most effectiveness in this mode.

Fine-tuning improves several models across both modes, notably boosting the Success Rate and ROR of all fine-tuned open-source models. Fine-tuning notably increased the Success Rate (SR) for models like LLaMA3.1-8B and Qwen2-7B, raising their SR from 2.17 to 23.91 and 4.35 to 19.57, respectively. The Reasonable Operation Ratio (ROR) also saw improvements, with models such as CogVLM2 jumping from 17.97 to 85.58 after fine-tuning.

\subsection{Additional Findings}

\vpara{Influence of Instruction Tuning.} Instruction tuning significantly enhances the performance of models across all four metrics in both XML and SoM modes, lifting the average success rates from 4.59\% to 21.50\% for LLMs and from 1.93\% to 13.28\% for LMMs. Notably, GLM4-9B's success rate rose to 21.01\%, with its Reasonable Operation Ratio (ROR) improving to 93.25, indicating better operational efficiency. The Reversed Redundancy Ratio (RRR) saw consistent gains, demonstrating reduced unnecessary actions, such as GLM4-9B improving its RRR from 54.43 to 74.81.

In SoM mode, models like CogVLM2, LLaMA3.2-11B, and Qwen2-VL-7B showed significant advancements across all four metrics. Qwen2-VL-7B's SR increased from 3.62 to 18.12\%, and its ROR rose to 88.29. The Sub-SR and RRR also benefited from tuning, marking improved task breakdown and reduced redundancy. After tuning, the best-performing open-source LLMs are approaching the level of GPT-4o, while the top LMMs have surpassed Gemini-1.5-Pro, reflecting comprehensive improvements across success, operational efficiency, and task execution. The tuned models' effective actions (ROR) have also surpassed those of most closed-source models, demonstrating enhanced precision. 

\begin{figure}[t]
    \centering
    \includegraphics[width=220pt]{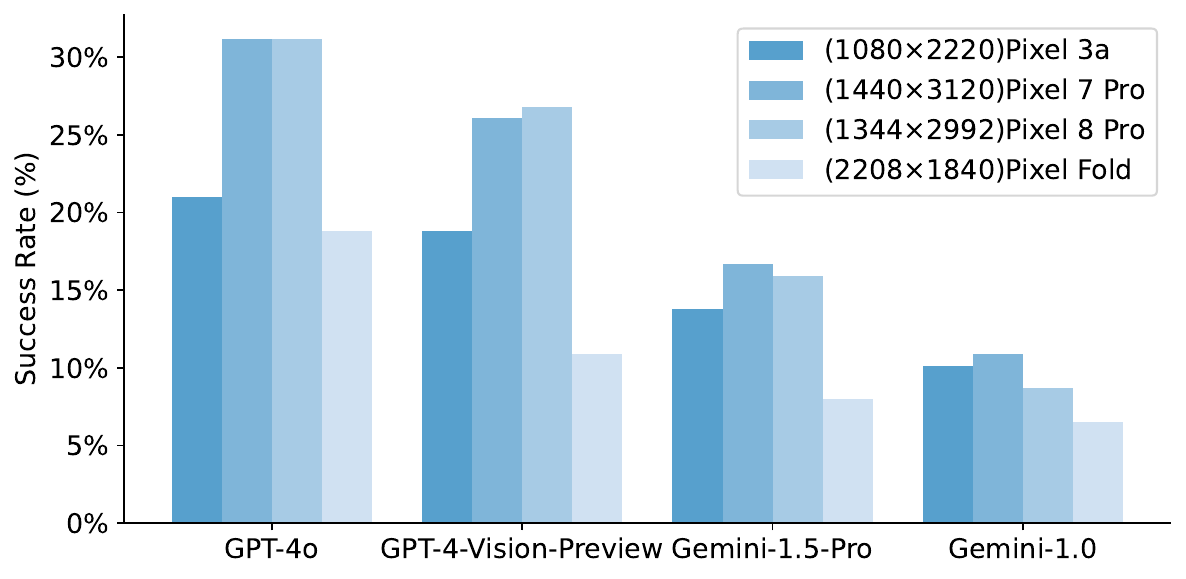}
    \caption{The performance of four models across four different device types is presented. Among these, the Pixel 3a is a smaller-sized phone, the Pixel 7 Pro and Pixel 8 Pro are of sizes comparable to commonly used phones, and the Pixel Fold is akin to a tablet.}
    \label{fig:diff Windows Size}
\end{figure}

\vpara{Influence of Windows Size.} As shown in Figure~\ref{fig:diff Windows Size}, experiments with three Android VMs of varying sizes in SoM mode show optimal agent performance on screens matching commonly used smartphones (e.g., Pixel 7 Pro, Pixel 8 Pro). Performance drops on smaller (Pixel 3a) and larger screens (Pixel Fold) due to increased scrolling needs and landscape orientation challenges, respectively.


\begin{table}[h!]
    \renewcommand{\arraystretch}{0.7}
    \centering
    \caption{The impact of the ReAct and SeeAct frameworks on SR results. Notably, model performance is significantly improved in XML+ReAct mode. Full results of this table are shown in Appendix~\ref{appendix:table-react-seeact-full}}
    \label{tab:table-react-seeact-sr}
    \begin{tabular}{@{}llc@{}}
    \toprule
    \textbf{Mode}     & \textbf{Model} & \textbf{SR} \\ \midrule
    \multirow{2}{*}{XML}        & GPT-4o                   & 25.36       \\
                                & Gemini-1.5-Pro           & 18.84       \\ \midrule
    \multirow{2}{*}{XML+ReAct}  & GPT-4o                   & 33.33       \\
                                & Gemini-1.5-Pro           & 31.16       \\ \midrule
    \multirow{2}{*}{XML+SeeAct} & GPT-4o                   & 24.64       \\
                                & Gemini-1.5-Pro           & 21.01       \\ \midrule
    \multirow{2}{*}{SoM}        & GPT-4o                   & 31.16       \\
                                & Gemini-1.5-Pro           & 16.67       \\ \midrule
    \multirow{2}{*}{SoM+ReAct}  & GPT-4o                   & 31.88       \\
                                & Gemini-1.5-Pro           & 15.94       \\ \midrule
    \multirow{2}{*}{SoM+SeeAct} & GPT-4o                   & 30.43       \\
                                & Gemini-1.5-Pro           & 21.01       \\ \bottomrule 
    \end{tabular}
\end{table}

\begin{table}[h!]

    \centering
    \caption{Average generation tokens of different modes. We used the LLaMA3 tokenizer for calculation. FT represents instruction tuning models.}
    \label{tab:tokens}
    \resizebox{0.45\textwidth}{!}{
    \begin{tabular}{@{}lcccc@{}}
    \toprule
    \textbf{Mode} & FT & XML/SoM & ReAct & SeeAct \\ \midrule
    \#Avg. Gen. Tokens & 4.96 & 23.56 & 67.89 & 129.12 \\ \bottomrule
    \end{tabular}}
\end{table}

\vpara{Analysis of Agent Frameworks.} We assess ReAct and SeeAct frameworks with GPT-4o and Gemini-1.5-Pro in XML and SoM modes. Table\ref{tab:table-react-seeact-sr} shows ReAct significantly improves performance only in XML mode. SeeAct does not enhance performance consistently due to the model's reasoning limitations with multimodal input. We also compare the SoM framework and bbox-only and show SoM is better; please refer to Appendix~\ref{tab:som vs bbox} for more detail.
ReAct and SeeAct frameworks increase token usage, harming efficiency. As per Table~\ref{tab:tokens}, XML+ReAct settings produce an average of 67.89 tokens, while models post-Instruction Tuning averaged only 4.96 tokens.

\section{Conclusion}
In this paper, we introduced \textsc{AndroidLab}, which includes a standard operational environment and a benchmark for agents interacting with Android devices. By integrating the XML and SoM operation modes, we ensured that the action space was consistent, enabling fair comparisons across different models. \textsc{AndroidLab} benchmark encompasses 138 tasks from nine apps, focusing on reproducibility and real-world relevance, allowing for precise task completion and progress assessment. We also introduced the Android Instruct dataset, comprising 10.5k traces and 94.3k steps, which significantly boosted the performance of open-source models when used for fine-tuning.

Our experiments demonstrated that fine-tuned open-source models have shown considerable improvements while top-performing closed-source models like GPT-4o and Claude-3.5-Sonnet  continue to lead in success rates and efficiency. Notably, fine-tuning raised success rates and operational efficiency, helping some models approach or even surpass closed-source counterparts in certain metrics. These findings highlight the potential of open-source models to enhance mobile agent performance, suggesting that further fine-tuning and optimization could narrow the gap between open and closed-source solutions. Future work could explore minimizing redundancy and improving task efficiency, enhancing the practical deployability of Android agents.

\section*{Acknowledgment}
We would like to thank Zhipu AI for sponsoring the computation resources and annotation costs used in this work.

\clearpage

\bibliography{ref}
\bibliographystyle{acl_natbib}

\appendix

\section{Details of Tasks}
\label{appendix:all task}

In our experiment, we use various apps to conduct various tests (succinctly presented in Table~\ref{table:task-table}). The following mobile apps are chosen:
\begin{itemize}[leftmargin=*,itemsep=0pt,parsep=0.2em,topsep=0.2em,partopsep=0.0em]
    \item \textbf{Bluecoins}: A personal finance management app used for tracking expenses and income. 
    \item \textbf{Calendar}: A calendar app helps in organizing schedules and setting reminders. 
    \item \textbf{Cantook}: An e-book reader for storing, managing, and reading e-books. 
    \item \textbf{Clock}: A clock app for displaying the time, setting alarms, and using a stopwatch. 
    \item \textbf{Contacts}: A contact management app for storing and organizing contact information. 
    \item \textbf{Maps.me}: An offline map app for navigation and exploring locations. 
    \item \textbf{PiMusic}: A music player app for organizing and playing locally stored music files. 
    \item \textbf{Settings}: A settings app for configuring device settings and preferences. 
    \item \textbf{Zoom}: A video conferencing app for hosting and joining online meetings.
\end{itemize}
The selection of these apps goes through multiple iterations to ensure their suitability for our evaluation purposes. A key criterion for the final selection is that each app functions independently, without requiring an internet connection or user account login. This ensures that the evaluations can be consistently replicated under the same conditions, eliminating external dependencies and reducing the risk of privacy breaches. As a result, this approach maintains the reliability and reproducibility of our results. 

\begin{table*}[]
\caption{List of Android Eval apps used along with corresponding example task, sub-goals, and the number of tasks.}
\begin{tabularx}{\textwidth}{@{}lllc@{}}
\toprule
\textbf{APP} & \textbf{Example Task} & \textbf{Sub-Goals} & \textbf{\# tasks} \\ \midrule

Bluecoins & 
\begin{tabular}[c]{@{}l@{}}
Record an income of 8000 CNY in \\
the books, and mark it as "salary".
\end{tabular} & 
\begin{tabular}[c]{@{}l@{}}
· type: income \\ 
· cash: 8000 CNY \\
· note: salary
\end{tabular} &
15 \\ 
\midrule

Calendar  &
\begin{tabular}[c]{@{}l@{}}
Edit the event with title "work", \\ 
change the time to be 7:00 PM.
\end{tabular} & 
\begin{tabular}[c]{@{}l@{}}
· title: work \\
· state: editing \\
· date: today \\
· time: 7 PM
\end{tabular} &
14 \\ 
\midrule

Cantook &
\begin{tabular}[c]{@{}l@{}}
Mark Hamlet as read.
\end{tabular} &
\begin{tabular}[c]{@{}l@{}}
· book: Hamlet \\
· state: 100\% read
\end{tabular} &
12 \\
\midrule

Clock & 
\begin{tabular}[c]{@{}l@{}}
I need set an 10:30PM clock every \\
weekend, and label it as "Watch \\
Football Games".
\end{tabular} &
\begin{tabular}[c]{@{}l@{}}
· time: 10:30PM \\
· frequency: every weekend \\
· label: Watch Football Games
\end{tabular} & 
27 \\ 
\midrule

Contacts &
\begin{tabular}[c]{@{}l@{}}
Add a contacts whose name is Xu, \\
set the working phone number to be \\
12345678, and mobile phone num- \\
ber to be 87654321.
\end{tabular} &
\begin{tabular}[c]{@{}l@{}}
· name: Xu\\
· working phone number: 12345678\\
· mobile phone number: 87654321
\end{tabular} &
15 \\
\midrule

Maps.me &
\begin{tabular}[c]{@{}l@{}}
Check the driving distance and time \\
between Bus stop of 2700 Coast Av- \\
enue and Bus Stop Route 51.
\end{tabular} &
\begin{tabular}[c]{@{}l@{}}
· driving distance: 7.0km \\
· driving time: 8 min
\end{tabular} &
15 \\ 
\midrule

PiMusic &
\begin{tabular}[c]{@{}l@{}}
Sort Pink Floyd's songs by duration \\
time in descending order.
\end{tabular} &
\begin{tabular}[c]{@{}l@{}}
· page: ARTISTS \\
· artist: Pink Floyd \\
· order: descending by duration
\end{tabular} &
12 \\
\midrule

Setting &
\begin{tabular}[c]{@{}l@{}}
Show battery percentage in status \\
bar.
\end{tabular} &
\begin{tabular}[c]{@{}l@{}}
· battery percentage: displayed 
\end{tabular} &
23 \\ 
\midrule

Zoom &
\begin{tabular}[c]{@{}l@{}}
I need to join meeting 1234567890 \\
without audio and video.
\end{tabular} &
\begin{tabular}[c]{@{}l@{}}
· meeting ID: 1234567890 \\
· audio: off \\
· video: off
\end{tabular} &
5 \\

\arrayrulecolor{black}\bottomrule
\end{tabularx}
\label{table:task-table}
\end{table*}

\section{Detail of Operation Modes}
\label{appendix:mode}
\subsection{XML mode}

\label{sec:xml-mode}

\begin{figure*}
    \centering
    \includegraphics[width=1\linewidth]{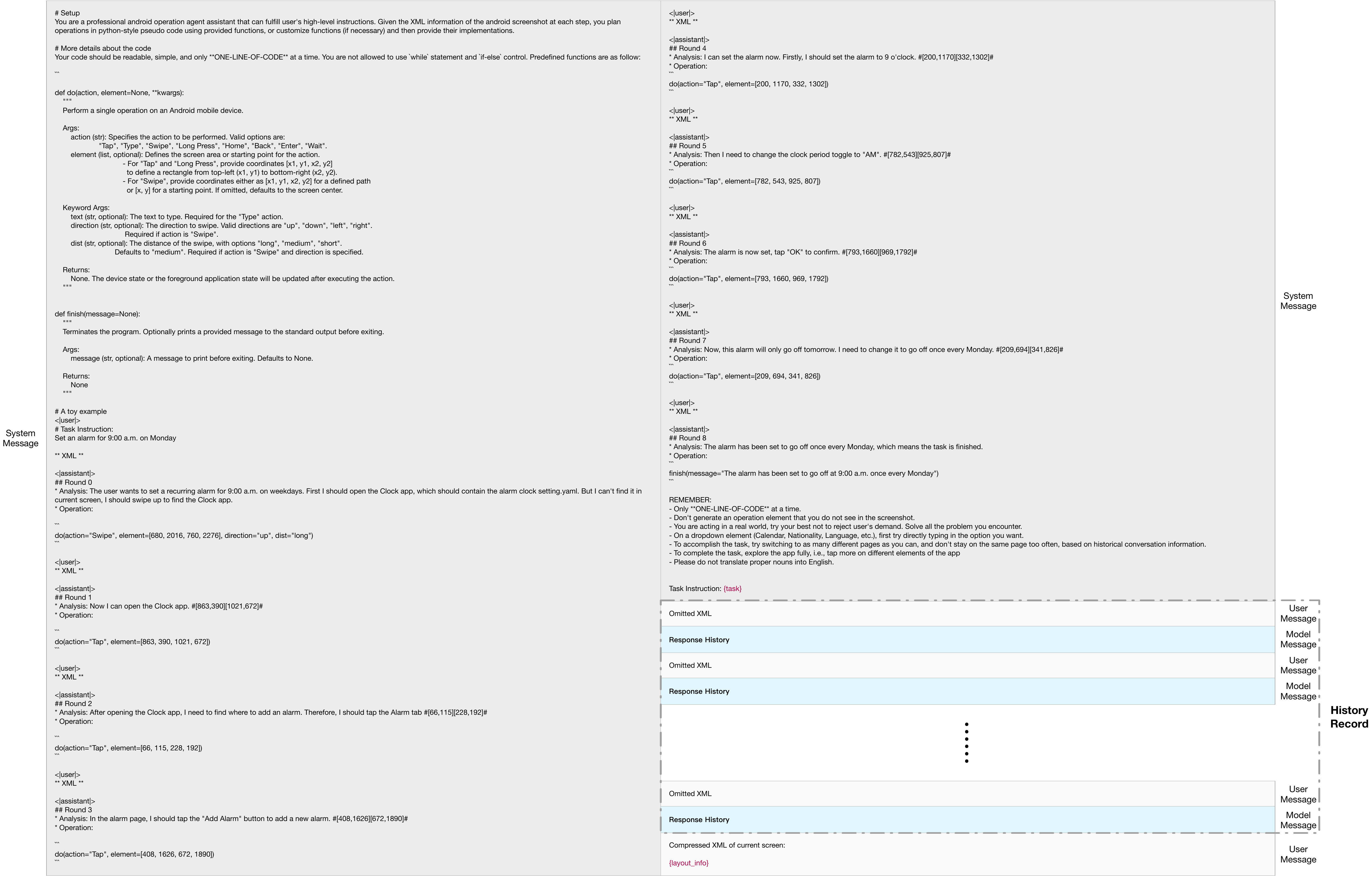}
    \caption{Prompts of XML Mode for Text-only Testing}
    \label{fig:naive-text}
\end{figure*}

As shown in Figure \ref{fig:naive-text}, in this mode, we prompt models with a task description, interaction history, and current compressed XML information. The models are supposed to output an action in function-call format. The actions are applied on coordinates shown in XML.

\subsection{SoM mode}

\label{sec:som-mode}

\begin{figure*}
    \centering
    \includegraphics[width=1\linewidth]{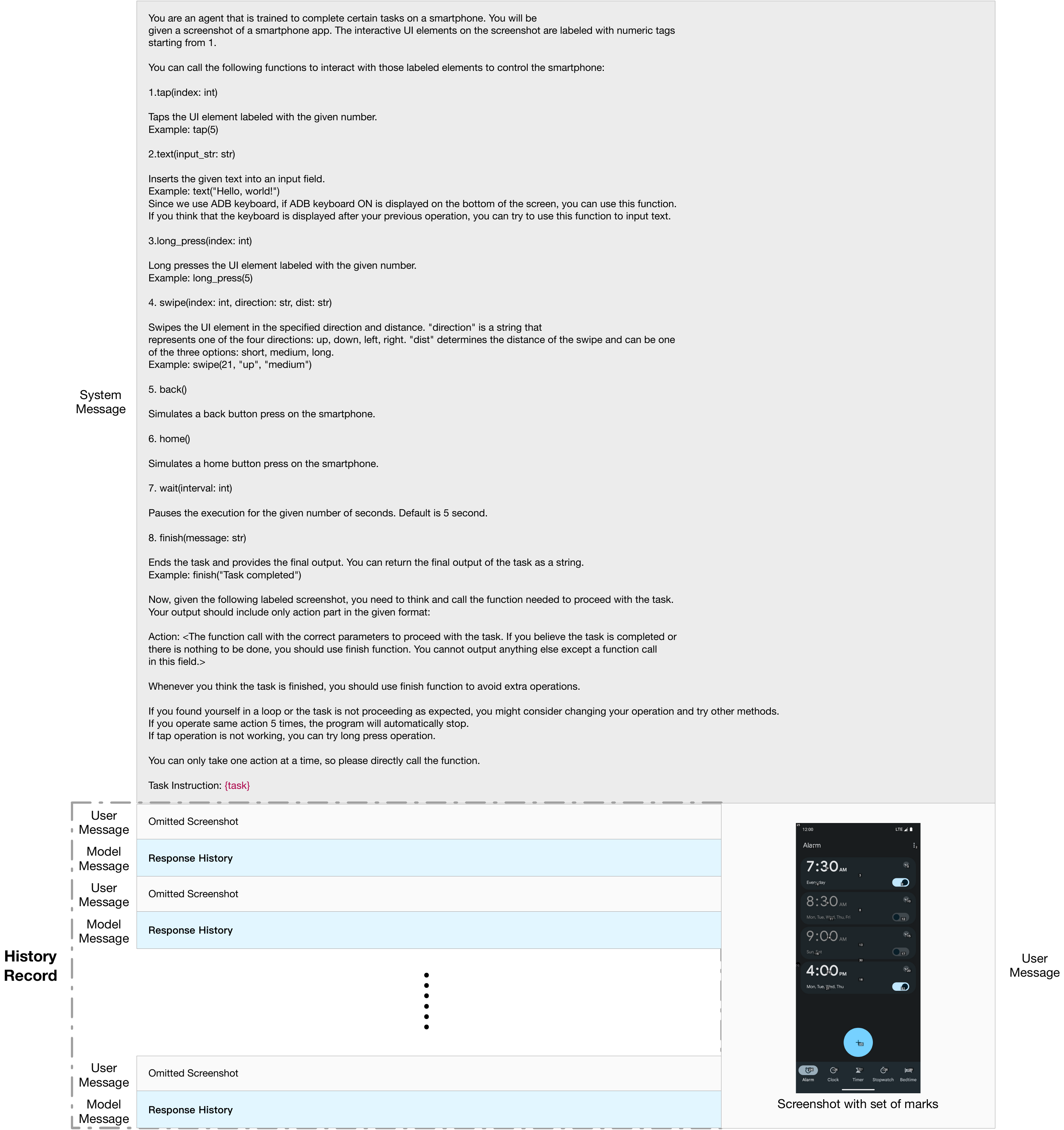}
    \caption{Prompts of SoM Mode for Multi-modal Testing}
    \label{fig:naive-screen}
\end{figure*}

As shown in Figure \ref{fig:naive-screen}, in this mode, we prompt models with a task description, interaction history, and current screenshot with a set of marks\cite{yang2023setofmark}. The models are also supposed to output an action in function-call format. Different from XML mode, the actions are performed on specified elements via marked indices.

\subsection{ReAct mode}

We follow \cite{yao2022react} for ReAct prompting. In this mode, we perform both text-only and multi-modal testing.
The text-only and multi-modal prompts are based on Section \ref{sec:xml-mode} and Section \ref{sec:som-mode} respectively. We both add prompts that allow models to think step by step before output actions.

\subsection{SeeAct mode}

\begin{figure*}
    \centering
    \includegraphics[width=0.96\linewidth]{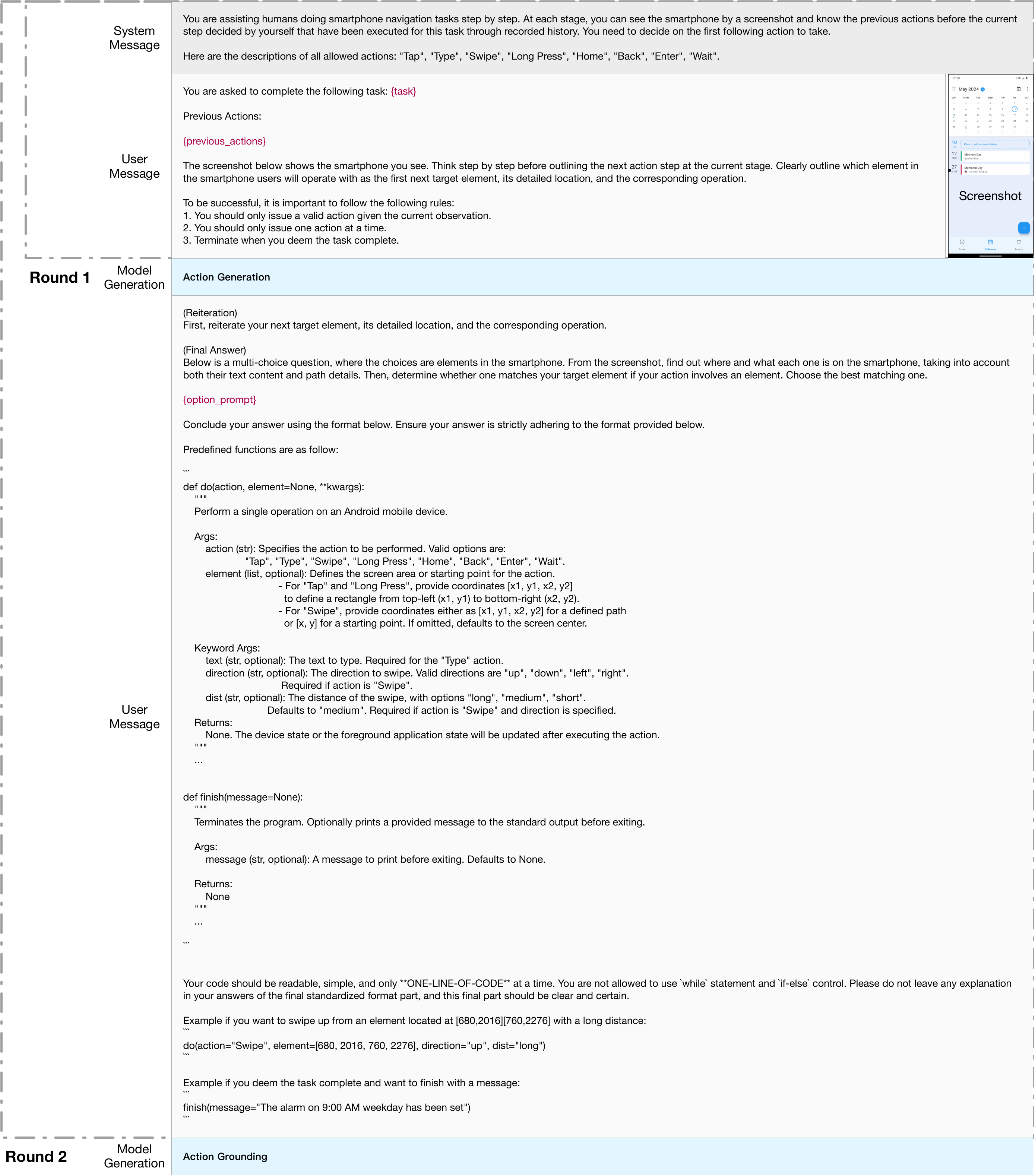}
    \caption{SeeAct Prompts for Multi-modal Testing}
    \label{fig:prompt-seeact-screenshot}
\end{figure*}

\begin{figure*}
    \centering
    \includegraphics[width=1\linewidth]{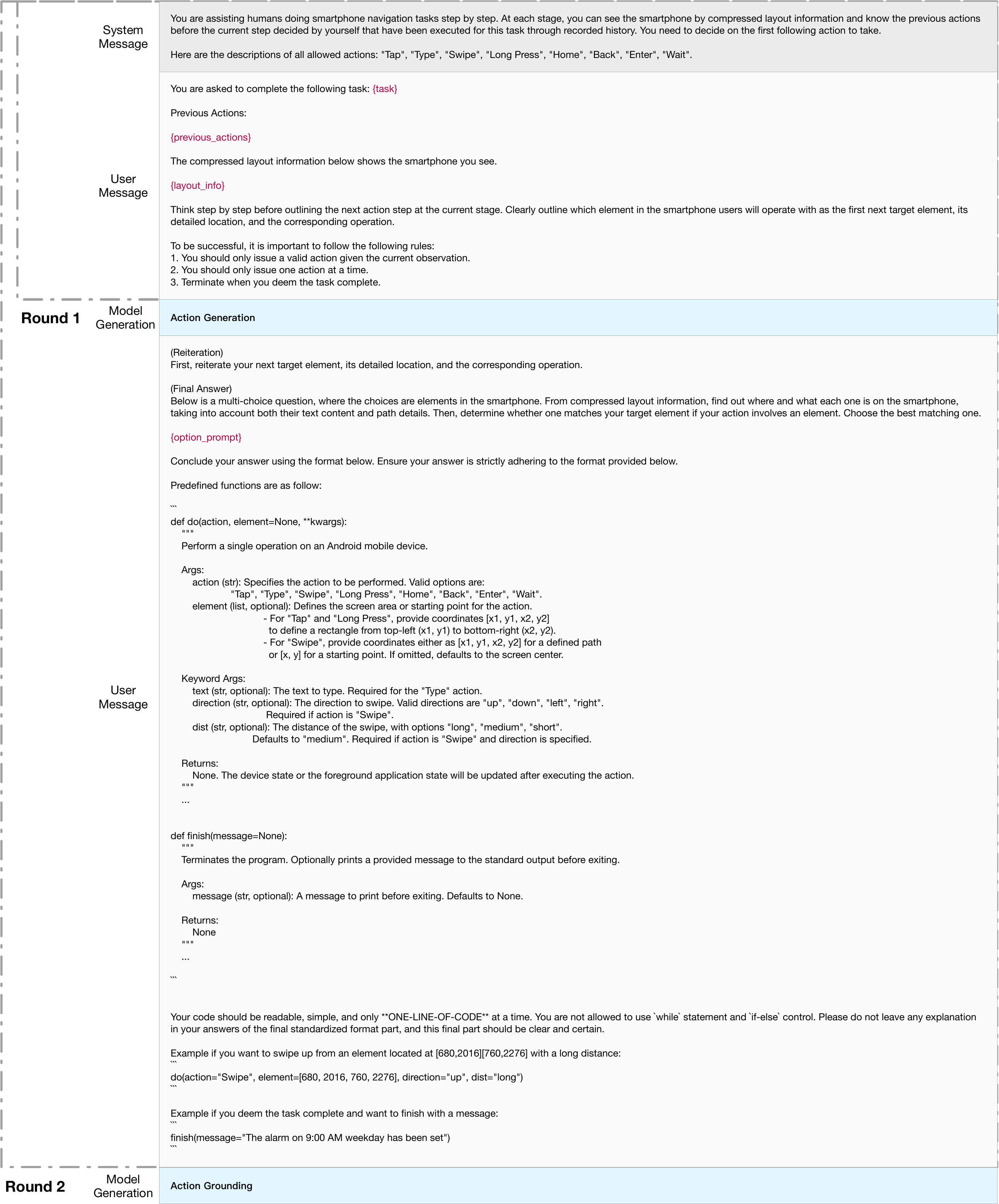}
    \caption{SeeAct Prompts for Text-only Testing}
    \label{fig:prompt-seeact-text}
\end{figure*}

We follow \cite{zheng2024gpt-seeact} for SeeAct prompting. The raw prompts of SeeAct are designed for web browsers. To adopt that in Android environments, we make some modifications, and the final prompts are shown in Figure \ref{fig:prompt-seeact-screenshot} for multi-modal testing and Figure \ref{fig:prompt-seeact-text} for text-only testing. 

For multi-modal and text-only testing, the information on mobile phones is given by screenshots and compressed XML respectively. The models are supposed to generate detailed description of the action and its corresponding element and parameters in round 1, and the expected function-call format in round 2.
\section{Details of Android Instruction Dataset}
\label{appendix:training data}

\subsection{Details of Human Annotation}

In the process of constructing our data, we utilize crowdsourced annotations. To ensure that the privacy information of the annotators is not disclosed, we adopt the following measures:

\begin{enumerate} \item Before the annotation begins, we explicitly inform the annotators that the annotated data will be used to fine-tune models, and part of the data will be open-sourced. Annotators who disagree may opt out of the annotation process. \item During the annotation process, all annotated data are first stored locally by the annotators. If an annotator believes that specific data involves privacy disclosure, they may choose not to use it or skip the task. \item After the annotation is completed, we mask and replace sensitive information such as usernames and chat logs before using the data for training. Additionally, such data will not be open-sourced. \end{enumerate}

All annotators sign formal contracts and are compensated according to reasonable standards.

\subsection{Instructions Given To Annotators}

We provide the instructions given to the annotators below. Note that our targets are expanded by hand-written instructions or academic datasets with available licenses.

\textbf{Task Overview}

For each labeling task, a target task will be given, such as: \textit{Navigate to XXX using Amap (Gaode Map)}.

The annotator must complete the task using their phone and follow the labeling process described below to ensure it is accurately executed and recorded.

To perform this annotation task, you must install ADB (Android Device Bridge) on your computer to control the phone and install the corresponding APK. Since the task involves collecting low-level information, we will require the phone to enable multiple permissions. Still, we guarantee that the information will not be transmitted in real-time during collection. The transmitted information includes the operation details, screenshots before and after each operation, and the corresponding XML files (only containing information from the current page). You can review and decide whether to keep the annotation data. If the annotation process involves screenshots or other information that you do not want to be used for training, you can:

\begin{enumerate}
    \item Skip the screenshot or specify that parts of the screenshot be hidden.
    \item Skip the entire target task.
    \item Skip all tasks involving the currently annotated app.
\end{enumerate}

Your data will not be used for purposes other than training the model.

After completing the annotation, you must upload all the tasks you were responsible for in one go. We have designed a plugin to store all the content in a unified folder.

A complete annotation consists of multiple operations called a sequence (trace). Each single-step operation is recorded once, and the definition of a single-step operation is detailed in the annotation documentation.

Please follow the steps below for plugin usage to install the annotation plugin.

\textbf{Plugin Usage Instructions}

\textit{Installing ADB and Connecting Phone to Computer}

For your Android phone, you need to perform the following settings:

\begin{enumerate}
    \item Connect the phone to the computer via a USB cable.
    \item Ensure that the \textbf{Developer Options} and \textbf{USB Debugging Mode} are enabled on the Android phone:
    \begin{itemize}
        \item Go to \textit{Settings} - \textit{Developer Options} - \textit{Android Debugging}. Check the box for \textit{Allow USB debugging}. If unavailable, go to \textit{Settings - System Updates - Developer Options - USB Debugging}.
        \item If you can't find the developer options, go to \textit{Settings} - \textit{About Phone} and tap the \textit{Build Number} seven times.
        \item If these methods don't work, search for how to enable developer options and USB debugging specific to your phone model.
        \item If you still encounter issues, seek help in the group chat.
    \end{itemize}
    \item Reconnect the phone to the computer, and on the phone, click \textit{Allow file transfer/USB debugging/higher permissions}. Also, allow the connection on the computer (if prompted).
    \item After entering Developer Mode, turn off the following animations under \textit{Developer Options} to increase the success rate of retrieving XML information via ADB commands:
    \begin{itemize}
        \item Window Animation Scale.
        \item Transition Animation Scale.
        \item Animator Duration Scale.
    \end{itemize}
\end{enumerate}

Follow the steps above until the following result is displayed using the command \textit{adb devices}:

\texttt{adb devices}

\texttt{List of devices attached}

\texttt{1a0d5d59    device}

The number before \textit{device} is randomly generated. You should see only one device. If there is more than one, try disconnecting other devices or closing virtual machines.

\textit{Installing ADB Keyboard}

Download the ADB Keyboard APK.

Run: \texttt{adb install <APK full path>}

Enable permissions on the phone and agree to the installation.

Once the installation is complete, set ADB Keyboard as the default input method in the phone settings. You can try the following two lines of code:

\begin{verbatim}
ime enable com.android.adbkeyboard/.AdbIME
ime set com.android.adbkeyboard/.AdbIME 
\end{verbatim}

If successful, when you open any text box, you'll see the message \textit{ADB Keyboard ON} at the bottom of the screen. If unsuccessful, manually change the input method in the settings.

\textit{Running Test Script}

\begin{enumerate}
    \item Open the command line, run \textit{adb devices}, and ensure correct output.
    \item Run the following commands in adb shell:
    \begin{verbatim}
    input keyevent KEYCODE_BACK
    input keyevent KEYCODE_HOME
    input keyevent KEYCODE_ENTER
    \end{verbatim}
    If there's no error or response, it's fine. If you see \textit{Command execution failed}, ensure you're using the correct method sequence, not \textit{Press xxx} commands like \textit{adb shell input keyevent KEYCODE\_A}.
    \item Open any text input field and run the following commands in adb shell:
    \begin{verbatim}
    input keyevent KEYCODE_A
    \end{verbatim}
    The setup succeeds if the letter "a" appears on the screen.
\end{enumerate}

\textbf{Annotation Plugin Usage Instructions}

You can perform the following operations on the phone. After completing any one of these operations, do not proceed until the command line shows \textit{Operation completed}. If the phone has not responded yet (such as loading a new page), wait until the page is fully loaded before clicking the next \textit{Begin}.

\begin{enumerate}
    \item \textbf{Click or Swipe}: Perform this directly on the phone. Click slowly, holding for 0.2 to 0.5 seconds. 
    \item \textbf{Text Input}: If the ADB Keyboard was successfully installed, you can input text. Before entering text, click on the text box in the previous step and ensure that the \textit{ADB Keyboard ON} symbol appears at the bottom of the screen. Click the \textit{Type} button on the GUI interface, enter the desired text in the computer's input box (Chinese/English), then click \textit{OK}. You will observe the input on the phone, and the command line will display \textit{Simulating typing xxx}.
    \item \textbf{Press xxx}: Three preset buttons are defined: \textit{Press Home} (Home key), \textit{Press Back} (Back key), and \textit{Press Enter} (keyboard Enter key). The command line will show \textit{Simulating press xxx}.
    \item \textbf{Finish Task}: If you believe the task is complete, click the \textit{Finish} button on the GUI. If the task requires an answer, fill in the response in the popup text box. If not, click \textit{OK}.
\end{enumerate}

After finishing a task, you can close the command line and GUI windows. If there are no issues with the annotation, you can return to Step 2 to start the next annotation. Otherwise, follow these steps:

\begin{enumerate}
    \item The command line will output the \textit{Save Path}, which contains all saved information for the annotation. You may delete the folder if you believe an error occurred or sensitive information was recorded.
    \item Each task has a prefix consisting of the first 32 characters of the task name. Ensure that the final submission includes one and only one instance of each non-skipped task.
    \item If certain operations were recorded incorrectly without affecting the phone's state, you may delete those steps. The step sequence is stored in \textit{Save Path/traces/trace.jsonl}. Record the steps you need to delete.
    \item If a screen contains sensitive information that can be removed while still being used for training, record the steps and describe the sensitive information in detail.
\end{enumerate}

\textbf{Summary of Key Points}

\begin{enumerate}
    \item Always use \textit{adb devices} before starting the annotation to ensure a successful connection.
    \item Reopen the app\_for\_xxx/dist/label(.exe) for each annotation instruction.
    \item The storage path must not contain Chinese characters.
    \item Click \textit{Begin} before each operation and wait for the message \textit{Begin your operation...} to appear before proceeding. If you proceed without waiting, the operation will be invalid. If the state cannot be recovered, you must restart the task. Make sure to click \textit{Begin} before finishing as well.
    \item After each operation is completed, wait until the corresponding success message appears in the command line and you see the output \textit{Operation completed} before clicking \textit{Begin} for the next action. Failure to follow these two key rules may result in invalid data. It’s better to proceed slowly and carefully than rush and make mistakes.
\end{enumerate}

\section{Additional Results}
\label{appendix:AR}
\begin{table*}[t]
    \caption{The number of tasks completed by all models across all apps in different modes.}
    \label{tab:apps-results}
    \resizebox{\textwidth}{!}{
    \begin{tabular}{@{}llcccccccccc@{}}
    \toprule
    \textbf{Mode}        & \textbf{Model}                          & \textbf{\begin{tabular}[c]{@{}c@{}}Bluecoins\\ 15\end{tabular}} & \textbf{\begin{tabular}[c]{@{}c@{}}Calendar\\ 14\end{tabular}} & \textbf{\begin{tabular}[c]{@{}c@{}}Cantook\\ 12\end{tabular}} & \textbf{\begin{tabular}[c]{@{}c@{}}Clock\\ 27\end{tabular}} & \textbf{\begin{tabular}[c]{@{}c@{}}Contacts\\ 15\end{tabular}} & \textbf{\begin{tabular}[c]{@{}c@{}}Maps.me\\ 15\end{tabular}} & \textbf{\begin{tabular}[c]{@{}c@{}}PiMusic\\ 12\end{tabular}} & \textbf{\begin{tabular}[c]{@{}c@{}}Setting\\ 23\end{tabular}} & \textbf{\begin{tabular}[c]{@{}c@{}}Zoom\\ 5\end{tabular}} & \textbf{\begin{tabular}[c]{@{}c@{}}Total\\ 138\end{tabular}} \\ \midrule

    \multirow{10}{*}{XML} & GPT-4o                                   & 1 & 0 & 3 & 8 & 5 & 5 & 2 & 10 & 1 & 35 \\
                          & GPT-4-1106-Preview                       & 1 & 4 & 6 & 4 & 6 & 6 & 4 & 9  & 3 & 43 \\
                          & Gemini-1.5-Pro                           & 1 & 1 & 3 & 6 & 3 & 4 & 3 & 4  & 1 & 26 \\
                          & Gemini-1.0                               & 0 & 1 & 1 & 4 & 2 & 0 & 1 & 2  & 1 & 12 \\
                          & GLM4-PLUS                                & 2 & 0 & 4 & 9 & 6 & 3 & 2 & 10  & 2 & 38 \\
                          & LLaMA3.1-8B-Instruct                     & 0 & 0 & 0 & 2 & 0 & 0 & 0 & 1  & 0 & 3  \\
                          & Qwen2.5-7B-Instruct                      & 0 & 0 & 2 & 1 & 1 & 0 & 0 & 2  & 0 & 6 \\
                          & GLM4-9B-Chat                             & 0 & 1 & 0 & 2 & 1 & 1 & 0 & 3  & 2 & 10 \\
                          & LLaMA3.1-8B-\textbf{ft}                  & 3 & 1 & 6 & 7 & 6 & 5 & 0 & 4  & 1 & 33 \\
                          & Qwen2.5-7B-\textbf{ft}                   & 1 & 1 & 3 & 4 & 7 & 4 & 1 & 6  & 0 & 27 \\
                          & GLM4-9B-\textbf{ft}                      & 0 & 1 & 5 & 7 & 5 & 2 & 0 & 8  & 1 & 29 \\ \midrule

    \multirow{11}{*}{SoM} & GPT-4o                                   & 1 & 1 & 5 & 7 & 8 & 2 & 2 & 13 & 4 & 43 \\
                          & GPT-4-Vision-Preview                     & 1 & 1 & 5 & 8 & 6 & 2 & 2 & 8  & 3 & 36 \\
                          & Gemini-1.5-Pro                           & 0 & 0 & 5 & 2 & 5 & 0 & 1 & 7  & 3 & 23 \\
                          & Gemini-1.0                               & 0 & 0 & 2 & 3 & 3 & 0 & 1 & 5  & 1 & 15 \\
                          & Claude-3.5-Sonnet                        & 4 & 2 & 4 & 9 & 7 & 0 & 3 & 10 & 1 & 40 \\
                          & Claude-3-Opus                            & 1 & 0 & 1 & 2 & 4 & 0 & 3 & 7  & 0 & 18 \\
                          & CogVLM2                                  & 0 & 0 & 0 & 0 & 0 & 0 & 0 & 1  & 0 & 1  \\
                          & LLaMA3.2-11B-Vision-Instruct             & 0 & 0 & 0 & 1 & 0 & 0 & 0 & 1  & 0 & 2  \\
                          & Qwen2-VL-7B-Instruct                     & 0 & 0 & 0 & 2 & 1 & 0 & 0 & 1  & 1 & 5  \\
                          & CogVLM2-\textbf{ft}                      & 0 & 0 & 2 & 3 & 4 & 1 & 1 & 4  & 1 & 16 \\
                          & LLaMA3.2-11B-Vision-\textbf{ft}          & 1 & 1 & 1 & 3 & 0 & 6 & 0 & 2  & 0 & 14 \\
                          & Qwen2-VL-7B-Instruct-\textbf{ft}         & 1 & 0 & 1 & 4 & 5 & 3 & 2 & 7  & 2 & 25 \\
                   \bottomrule
    \end{tabular}}
    \end{table*}

\begin{table*}[]
\caption{The improvement in model performance after employing the ReAct and SeeAct frameworks, is reflected in the increased number of successfully completed tasks across various apps.}
\label{tab:app-react-seeact}
\resizebox{\textwidth}{!}{
\begin{tabular}{@{}llcccccccccc@{}}
\toprule
\textbf{Mode} &
  \textbf{Model} &
  \multicolumn{1}{c}{\textbf{\begin{tabular}[c]{@{}c@{}}Bluecoins\\ 15\end{tabular}}} &
  \multicolumn{1}{c}{\textbf{\begin{tabular}[c]{@{}c@{}}Calender\\ 14\end{tabular}}} &
  \multicolumn{1}{c}{\textbf{\begin{tabular}[c]{@{}c@{}}Cantook\\ 12\end{tabular}}} &
  \multicolumn{1}{c}{\textbf{\begin{tabular}[c]{@{}c@{}}Clock\\ 27\end{tabular}}} &
  \multicolumn{1}{c}{\textbf{\begin{tabular}[c]{@{}c@{}}Contacts\\ 15\end{tabular}}} &
  \multicolumn{1}{c}{\textbf{\begin{tabular}[c]{@{}c@{}}Maps.me\\ 15\end{tabular}}} &
  \multicolumn{1}{c}{\textbf{\begin{tabular}[c]{@{}c@{}}PiMusic\\ 12\end{tabular}}} &
  \multicolumn{1}{c}{\textbf{\begin{tabular}[c]{@{}c@{}}Settings\\ 23\end{tabular}}} &
  \multicolumn{1}{c}{\textbf{\begin{tabular}[c]{@{}c@{}}Zoom\\ 5\end{tabular}}} &
  \multicolumn{1}{c}{\textbf{\begin{tabular}[c]{@{}c@{}}Total\\ 138\end{tabular}}} \\ \midrule
\multirow{2}{*}{XML}        & GPT-4o         & 1 & 0 & 3 & 8  & 5 & 5 & 2 & 10 & 1  & 35 \\
                            & Gemini-1.5-Pro & 1 & 1 & 3 & 6  & 3 & 4 & 3 & 4  & 1  & 26 \\ \midrule
\multirow{2}{*}{XML+ReAct}  & GPT-4o         & 2 & 0 & 4 & 12 & 7 & 6 & 2 & 11 & 2  & 46 \\
                            & Gemini-1.5-Pro & 4 & 0 & 4 & 6  & 6 & 6 & 3 & 11 & 3  & 43 \\ \midrule
\multirow{2}{*}{XML+SeeAct} & GPT-4o         & 1 & 2 & 4 & 8  & 5 & 3 & 2 & 7  & 2  & 34 \\
                            & Gemini-1.5-Pro & 1 & 0 & 6 & 6  & 5 & 0 & 2 & 8  & 1  & 29 \\ \midrule
\multirow{2}{*}{SoM}        & GPT-4o         & 1 & 1 & 5 & 7  & 8 & 2 & 2 & 13 & 4  & 43 \\
                            & Gemini-1.5-Pro & 0 & 0 & 5 & 2  & 5 & 0 & 1 & 7  & 3  & 23 \\ \midrule
\multirow{2}{*}{SoM+ReAct}  & GPT-4o         & 3 & 1 & 5 & 7  & 7 & 3 & 0 & 15 & 3  & 44 \\
                            & Gemini-1.5-Pro & 1 & 1 & 3 & 2  & 4 & 1 & 2 & 7  & 1  & 22 \\ \midrule
\multirow{2}{*}{SoM+SeeAct} & GPT-4o         & 6 & 1 & 4 & 11 & 6 & 0 & 2 & 9  & 3  & 42 \\
                            & Gemini-1.5-Pro & 1 & 0 & 6 & 6  & 5 & 0 & 2 & 8  & 1  & 29 \\ \bottomrule
\end{tabular}}
\end{table*}
\begin{table*}[]
\centering
\caption{Different multi-modal modes of instruction tuning. We use the same set of training data but only add a set-of-mask index on SoM mode. Note that AITW dataset even could not provide accurate bbox, but only point. We use CogVLM2 as base model.}
\label{tab:compare of it mode}
\begin{tabular}{ccccc}
\hline
\textbf{Operation Mode} & SR    & Sub-SR & RRR   & ROR   \\ \hline
BBOX                    & 5.79  & 6.03   & 47.95 & 55.05 \\
SoM                     & 11.59 & 16.06  & 57.37 & 85.58 \\ \hline
\end{tabular}
\end{table*}
\begin{table*}[h!]
    \renewcommand{\arraystretch}{0.7}
        \centering
    \centering
    \caption{The impact of the ReAct and SeeAct frameworks. Notably, model performance is significantly improved in XML+ReAct mode.
    }
    \label{tab:table-react-seeact-full}
    \begin{tabular}{@{}llcccc@{}}
    \toprule
    \textbf{Mode}     & \textbf{Model} & \textbf{SR} & \textbf{Sub-SR} & \textbf{RRR} & \textbf{ROR} \\ \midrule
    \multirow{2}{*}{XML}        & GPT-4o                   & 25.36       & 30.56           & 107.45       & 86.56        \\
                                & Gemini-1.5-Pro           & 18.84       & 22.40           & 57.72        & 83.99        \\ \midrule
    \multirow{2}{*}{XML+ReAct}  & GPT-4o                   & 33.33       & 38.22           & 97.93        & 90.74        \\
                                & Gemini-1.5-Pro           & 31.16       & 34.54           & 92.08        & 90.31        \\ \midrule
    \multirow{2}{*}{XML+SeeAct} & GPT-4o                   & 24.64       & 27.31           & 93.78        & 79.62        \\
                                & Gemini-1.5-Pro           & 21.01       & 25.53           & 75.97        & 89.06        \\ \midrule
    \multirow{2}{*}{SoM}        & GPT-4o                   & 31.16       & 35.02           & 87.32        & 85.36        \\
                                & Gemini-1.5-Pro           & 16.67       & 18.48           & 105.95       & 91.52        \\ \midrule
    \multirow{2}{*}{SoM+ReAct}  & GPT-4o                   & 31.88       & 39.19           & 104.69       & 89.80        \\
                                & Gemini-1.5-Pro           & 15.94       & 21.38           & 109.81       & 84.16        \\ \midrule
    \multirow{2}{*}{SoM+SeeAct} & GPT-4o                   & 30.43       & 36.24           & 97.45        & 88.56        \\
                                & Gemini-1.5-Pro           & 21.01       & 25.53           & 75.97        & 89.06        \\ \bottomrule 
    \end{tabular}
    \end{table*}
\subsection{Detail results across different APPs}
Table~\ref{tab:apps-results} shows the number of tasks correctly completed by various models across different apps, without employing the ReAct and SeeAct frameworks. This table shows that GPT-4o and GPT-4-1106-Preview perform relatively well, completing 78 and 79 tasks respectively. In the XML mode, GPT-4-1106-Preview stands out as the top performer with 43 tasks completed. Comparatively, in the SoM mode, GPT-4o excels, completing a significantly higher number of tasks than the other models. Most models exhibit high success rates in tasks like "Contacts" and "Setting". Overall, GPT-4o and GPT-4-1106-Preview outperform the other models significantly in both XML and SoM modes, while Gemini-1.5-Pro shows a reasonable number of task completions across various apps.

Table~\ref{tab:app-react-seeact} shows the performance improvements observed after implementing the ReAct and SeeAct frameworks on different models across various apps. Notably, GPT-4o shows significant enhancement, with the number of completed tasks increasing from 35 to 46 in XML+ReAct mode and from 43 to 44 in SoM+ReAct mode. Gemini-1.5-Pro also benefits, increasing from 26 to 43 tasks. The improvements are evident in specific apps like "Bluecoins", and are especially notable in high-complexity, multi-step tasks. GPT-4o leads in performance across all frameworks, showing how ReAct and SeeAct improve the model.

\subsection{Detail results across different multi-modal training mode}
\label{tab:som vs bbox}

We provide a comparison of different multimodal training modes in Table~\ref{tab:compare of it mode}. Under the same training data and base model settings, BBOX mode removes specified sets-of-masks from the screen. It is worth mentioning that datasets like AITW only provide click positions rather than bounding boxes (BBOX), and they do not offer a way to reconstruct the click-box from XML. Therefore, theoretically, data from AITW and similar datasets are more challenging to learn from.

\subsection{Detail results of SeeAct and ReAct methods}
\label{appendix:table-react-seeact-full}

We have provided detailed results on the impact of the SeeAct and ReAct frameworks on model performance in Fig~\ref{tab:table-react-seeact-full}, including all four metrics.

\end{document}